\documentclass[letterpaper, 10 pt, conference]{ieeeconf}  

\usepackage{graphicx}
\usepackage{epsfig}
\usepackage{bm}
\usepackage{amsmath}
\usepackage{amsfonts}
\usepackage{amssymb}
\usepackage{ascmac}
\usepackage{prettyref}
\usepackage{comment}
\usepackage{multirow}
\usepackage{makeidx}
\usepackage{fancyhdr}
\usepackage{dotseqn}
\usepackage[caption=false,font=footnotesize]{subfig}
\usepackage{float}
\usepackage{cite}
\usepackage{algpseudocode,algorithm}

\newrefformat{sec}{\ref{#1}“\nameref{#1}”}
\newrefformat{sub}{\ref{#1}“\nameref{#1}”}
\newrefformat{par}{\ref{#1}“\nameref{#1}”}
\newrefformat{tab}{\ref{#1}“\nameref{#1}”}
\newrefformat{fig}{Fig. \ref{#1}}

\newcommand{\argmax}{\mathop{\rm argmax}\limits}

\newcommand{\maxn}{\mathop{\rm max}\limits}

\IEEEoverridecommandlockouts                              

\title{\LARGE \bf
Assembly robots with optimized control stiffness \\ through reinforcement learning
}

\author{Masahide Oikawa$^{1}$, Kyo Kutsuzawa$^{1}$, Sho Sakaino$^{2}$ and Toshiaki Tsuji$^{1}$
\thanks{*This work was not supported by any organization}
\thanks{$^{1}$Masahide Oikawa, Kyo Kutsuzawa and Toshiaki Tsuji are with the Dept. Electrical and Electronic Systems, Saitama University, 255 Shimo-okubo, Saitama, 338-8570 
{\tt\small E-mail: tsuji@ees.saitama-u.ac.jp}}%
\thanks{$^{2}$Sho Sakaino is with the Graduate School of Systems and
Information Engineering, University of Tsukuba, 1-1-1 Tennodai,
Tsukuba, Ibaraki 305-8577, Japan and the JST PRESTO}%
}

\begin{document}

\maketitle
\thispagestyle{empty}
\pagestyle{empty}

\begin{abstract}
There is an increased demand for task automation in robots. Contact-rich tasks, wherein 
multiple contact transitions occur in a series of operations, are extensively being studied 
to realize high accuracy. In this study, we propose a methodology that uses reinforcement 
learning (RL) to achieve high performance in robots for the execution of assembly tasks that 
require precise contact with objects without causing damage. The proposed method ensures 
the online generation of stiffness matrices that help improve the performance of local 
trajectory optimization. The method has an advantage of rapid response owing to short 
sampling time of the trajectory planning. The effectiveness of the method was verified via 
experiments involving two contact-rich tasks. The results indicate that the proposed method 
can be implemented in various contact-rich manipulations.
\end{abstract}

\section{INTRODUCTION}
Automation of tasks involving contact is currently in great demand; 
for instance, robots performing assembly tasks that involve contact with the objects 
to be assembled without damaging the parts. High-order assembly may involve multiple 
contact transitions in a series of operations and such tasks are termed as 
``contact-rich tasks''. Because the actions 
to be performed by the robots change at each contact transition, the trajectory plannig 
and controller design are complex. Moreover, assembly often requires fitting parts into 
small gaps, measuring tens of micrometers, thereby generating the issue of achieving a 
level of accuracy that exceeds the robot's own position accuracy.

Impedance control \cite{hogan1985impedance, salisbury1980active} and 
admittance control \cite{lawrence1987position, kazerooni1990stability} are known to be 
effective in achieving higher accuracy in the assembly process than the robot's position 
accuracy. Because contact-rich tasks require further consideration of contact transitions, 
hybrid control was introduced, which involved recognizing the contact transitions and 
switching control modes~\cite{khatib1987unified}. The complexity of designing 
control mode switching during contact transitions is an issue in contact-rich tasks, because 
controllers tend to be conservative as the robot motion becomes unstable due to 
mode switching. Moreover, the accurate modeling of robots and their contact with objects 
pose difficulties. To support both autonomy and performance in assembly robots, a robust 
technology is required that recognizes the contact states without an accurate model and 
determines the input.

Reinforcement learning (RL) is a technique where the agents optimize the unknown steps through 
interactions with the environment; it is a potential method that can solve the aforementioned 
issues \cite{levine2016end,deisenroth2013survey,kober2013reinforcement}. In humans, for 
manual assembly tasks, the contact states are recognized from the sensation of touching 
the object based on heuristics, and an appropriate trajectory is provided by considering 
the contact state. RL renders this process autonomous. Several studies 
have already been conducted on this technology 
\cite{inoue2017deep,vecerik2018practical,ren2018learning,lee2019icra,ren2019fast,schoettler2019deep,luo2019reinforcement}.
In these studies, the local trajectories were optimized through the outputs by the agents to 
the controller of information on the positions or forces in the form of actions. However, 
owing to the nature of RL algorithms, obtaining the output of actions in 
short cycles (under a few tens of milliseconds) is difficult. The command output cycle is an 
element that is directly linked to the performance of the control system; this is a fundamental 
problem in robots that implement RL. The objective of this study is to 
maintain high control performance by outputting the position and force commands in brief 
cycles in RL. We propose an RL technique as a 
solution where the agent outputs the stiffness matrices as actions and demonstrate that 
the performance of local trajectory optimization is fundamentally improved. We verify the 
usefulness of the proposed method via a verification experiment for two examples of 
high-accuracy contact-rich tasks: peg-in-hole task and gear insertion task.

\section{Related works}
Many studies have been conducted on automating assembly 
tasks using deep reinforcement learning. Owing to the need for physical interaction 
with the environment for the purpose of learning, the number of trials for learning 
must be limited. Sim-to-real reduced the number of trials in the real world by 
retraining the model in the real world for which learning was performed in a 
simulation \cite{liu2018progressive, rusu2016sim, peng2018sim}. 

However, for high-accuracy contact-rich tasks involving multipoint contacts with 
gaps of a few tens of micrometers, the sim-to-real technique is not a realistic option 
owing to the divergence of the contact phenomenon between the simulation and the 
real world. A technique is therefore required to acquire the motor skills to perform 
high-accuracy contact-rich tasks through a relatively low number of real-world trials. 
Inoue {\it et al}. used the long short-term memory (LSTM) scheme for learning two independent 
policies to perform a peg-in-hole task \cite{inoue2017deep}. Luo {\it et al}. performed the 
contact-rich task of assembling a gear without pre-defined heuristics through an 
RL technique that derived the appropriate force control command 
values from the position and force responses \cite{luo2019reinforcement}. 
The methods show good performance based on adaptive impedance behavior 
through the adjustment of position/force commands, whereas their sampling periods 
are over a few tens of milliseconds. This paper shows the 
advantage of the agent that outputs stiffness matrices as an action of the RL.

On the basis of the aforementioned background, there are two main contributions of this study. 
First, this study shows that the local trajectory optimization can be done against an external force without 
changing the reference trajectory by adjusting only the stiffness matrix. 
Second, it shows the advantage of selecting stiffness matrices as an action produced by the RL agent.  
A shorter sampling period for the admittance model generating the position command leads 
to a higher performance of tasks with contact motion. 
Remote center compliance, a method that utilizes mechanical compliance, is a good candidate for optimizing 
the local trajectories of assembly actions without time delays \cite{ang1995specifying,vrcc}. However, we apply admittance 
control instead, because stiffness matrices ensure easy responses to the changes in fast-changing 
environmental conditions and tasks. The method proposed in this paper can also be applied 
to techniques that use mechanical elements. 

\section{Reinforcement learning with admittance control}
In this section, we first state the problem of this study. 
Subsequently, the control architecture of the admittance control is described. 
The important feature of this study is that the stiffness matrices are selected 
from the discrete action space by the RL agent. Hence, 
a method to design a stiffness matrix for the discrete action space is described. 
Then, the learning architecture with a DQN is shown after 
the trajectory planning algorithm.  
 
\subsection{Problem statement}
This study deals with peg-in-hole and gear-insertion tasks as the target for high precision assembly. 
The peg-in-hole task can be divided into two phases: search phase and insertion phase~\cite{care}. 
Teeth alignment phase, another additional phase should be considered in the gear-insertion task. 

The robot places the peg center within the clearance region of the hole center during the search phase. 
We use a 6-axis force-torque sensor to detect the relative relationship between the peg and the hole. 
Since it is difficult to obtain a precise model of the physical interaction between these two, 
the RL has been an effective solution to detect the relative relationship~\cite{inoue2017deep}. 
The robot adjusts the orientation of the peg by the admittance control and pushes the peg into the hole 
during the insertion phase. The peg often gets stuck in the initial stage of the insertion in case 
the clearance is small and some errors in attitude angle exist. 
In case of gear-insertion task, the gear wheel need to be matched to other gear wheels 
by aligning the gear teeth after the insertion. 
This study shows that the gear aligning trajectory can be generated by the local trajectory 
modification using a stiffness matrix. 

Although there are both cases in practice: inserting a grasped peg into a fixed hole; and inserting a 
fixed peg into a hole on a grasped part, these two cases are considered with the 
same control architecture. This paper described the method with the first case, while it can 
be applied to the second case as shown in the experimental verification results. 
  
%
\subsection{Control architecture}
\begin{figure}[tb]
    \centering
    \includegraphics[width=8.4cm,pagebox=artbox]{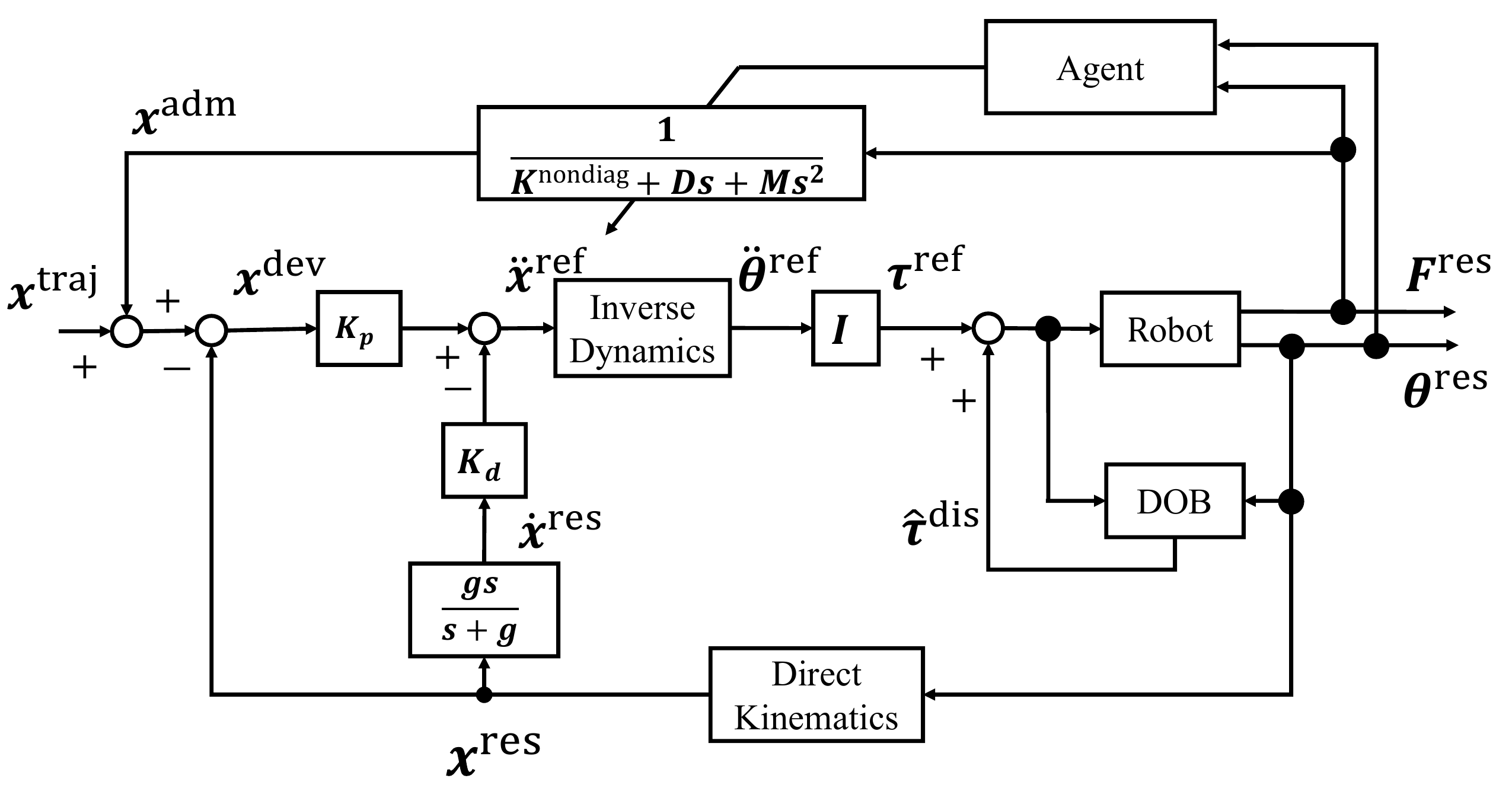}
    \caption{Block diagram of the admittance control}
    \label{fig:block_diagram_prop}
\end{figure}

Fig. \ref{fig:block_diagram_prop} shows a block diagram of the admittance control in this study. 
A disturbance observer (DOB) \cite{dob} was used in the control system to 
cancel the interference between the admittance model and PD controller. 
Note that admittance control based on a simple PD controller may not properly 
imitate the admittance model because PD controller itself work as a kind of 
a stiffness controller. 
Table \ref{tbl:parameter_control_real} lists the control parameters. We used a six 
degrees-of-freedom (six-DOF) serial-link manipulator. The system is an ordinary 
admittance control and the 
trajectory planner providing the trajectory input $\bm{x}^{traj}$ is also simple. 
The characteristic feature of the architecture is that the stiffness matrix 
$\bm{K}^{nondiag}$ is given by the agent. 

\begin{table}[htb]
    \centering
    \caption{Parameters of controller}
    \begin{tabular}{|l||l|l|} \hline
      $\bm{K_{p}}$  & Proportional gain & diag(500, 500, 500,  \\
       && 500, 500, 500) \\
      $\bm{K_{d}}$  & Derivative gain & diag(50, 50, 50,   \\
       && 50, 50, 50) \\
      $\bm{I}$  & MOI & diag(1.58, 2.08, 1.09,  \\
       && 0.081, 0.112, 0.035)  \\
       &&  [kgm$^{2}$] \\
      $g$  & Cutoff freq. of the derivative filter & 12 [Hz]\\
      $T^{c}_s$ & Sampling period of the controller& 0.001 [s]      \\
      $T^{NN}_s$  & Sampling period of the NN & 0.02 [s]      \\ \hline
    \end{tabular}
    \label{tbl:parameter_control_real}
\end{table}

\subsection{Trajectory modification using stiffness matrices}
This subsection outlines the technique used for the online generation of non-diagonal stiffness 
matrices using deep RL. 
The agent determines the action based on the state $\bm{s}_t$, as shown below.
\begin{eqnarray}
    \bm{s}_t= [p_x, p_y, p_z, f_x, f_y, f_z, \tau_{x}, \tau_{y}]
      \label{eqn:DQN_state}
\end{eqnarray}
where $p,f$ and $\tau$ denote tip position, external force, and external torque,
respectively. $x, y$ and $z$ in subscripts denote the axes.  
The action $\bm{a}_t$ selected by the agent is a non-diagonal stiffness matrix.
\begin{eqnarray}
    \bm{a}_t = \bm{K}^{nondiag}
      \label{eqn:DQN_action}
\end{eqnarray}

\begin{figure}[b]
    \centering
    \includegraphics[width=8cm]{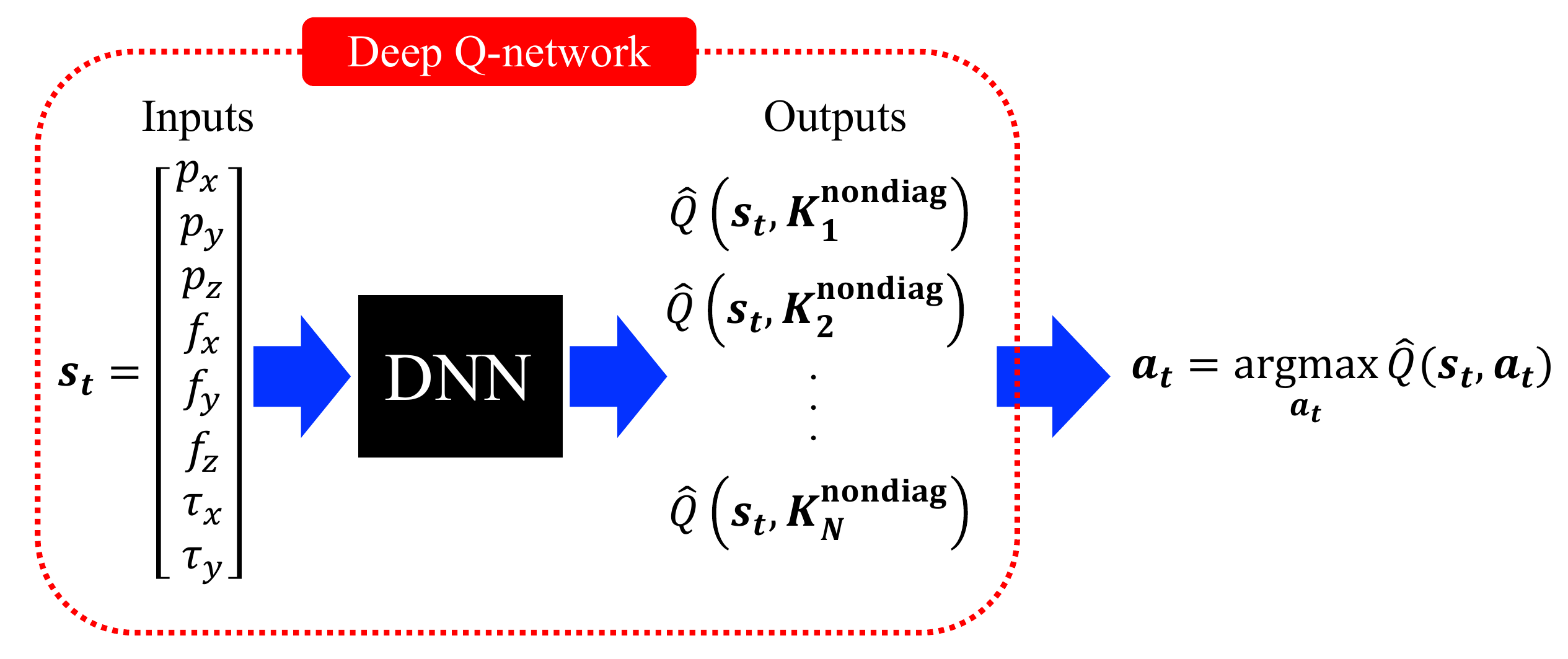}
    \caption{Action selected by the DQN}
    \label{fig:DQN_InOut_argmax}
\end{figure}

Generally, the outputs in Q-learning are discrete values. Thus, the output is the highest 
action value out of the few non-diagonal stiffness matrices designed in advance. 
The number and type of stiffness matrices to be designed vary depending on the task. 
\prettyref{fig:DQN_InOut_argmax} shows the inputs/outputs of the DQN 
and the flow of input into the robot. The current robot state $\bm{s}_t$ is input into the DQN 
and the value for each action (stiffness matrix $\bm{K}^{nondiag}$) is obtained as the output. 
The stiffness matrix with the highest action value obtained as the output of the DQN is the 
output of the robot and the stiffness matrix $\bm{K}^{nondiag}$ in the admittance control 
is updated. 
As discussed in \cite{kaneko}, the robot motion is made relative to the external force by adapting 
the non-diagonal terms of the stiffness matrices. This indicates that, 
while the trajectories are actively modified by the agent in the conventional 
techniques, they can be passively modified based 
on the external force in the proposed methodology. While the trajectory is modified 
in both techniques, the introduction of a passive methodology improves the responsiveness. 
The basis for this is described as follows. 

\begin{figure}[tb]
    \centering
    \includegraphics[width=8cm]{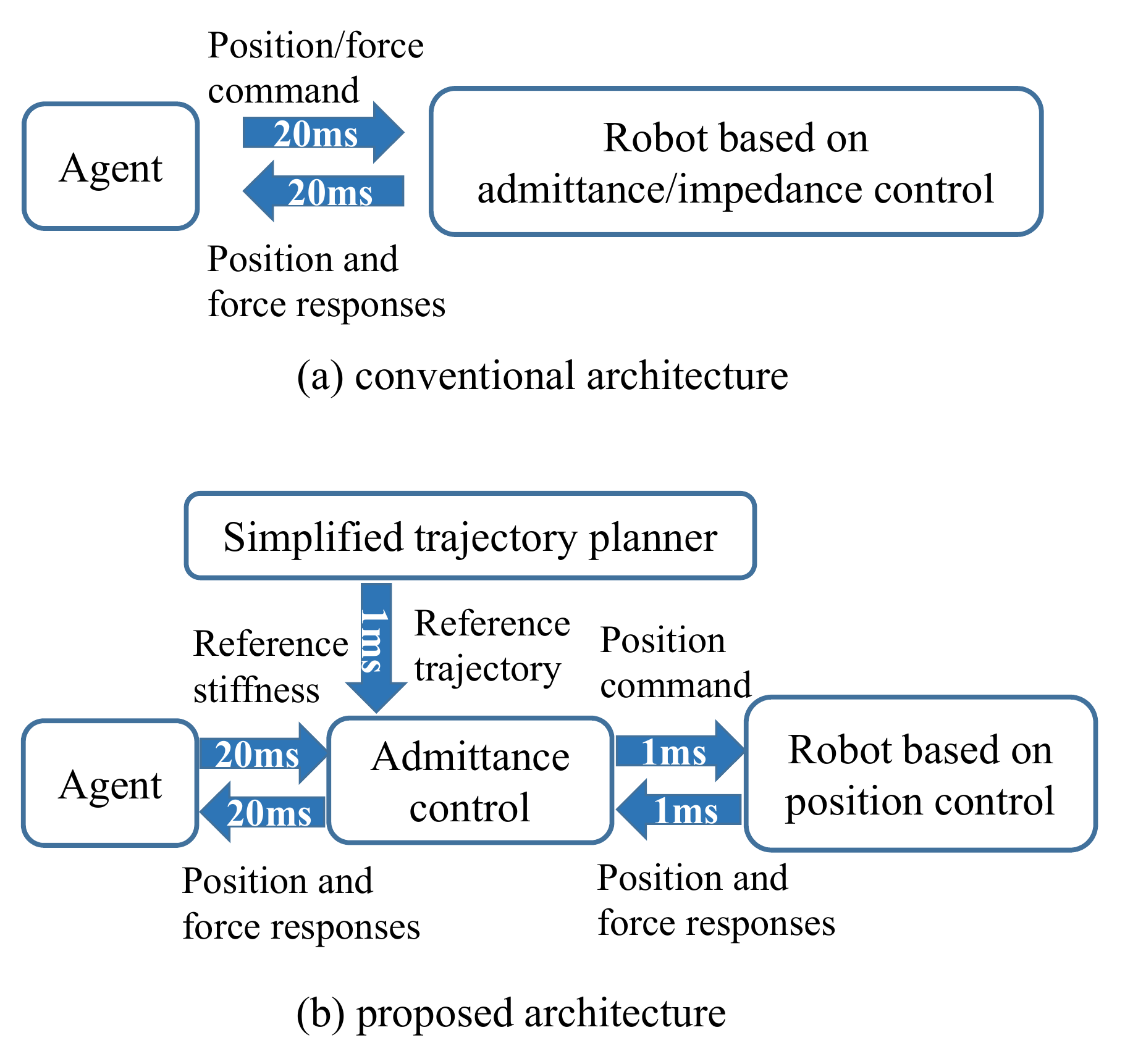}
    \caption{Architecture of the proposed method}
    \label{fig:architecture}
\end{figure}

The diagrams for action generation considering the contact states for both the proposed and 
conventional methods are compared in \prettyref{fig:architecture}. 
Generally, the sampling cycle of the agent is longer than the control cycle of the robot. 
In the proposed method, the agent generates the reference stiffness $\bm{K}^{nondiag}$ 
in cycles of 20 ms and the admittance controller generates a position command based 
on the reference trajectory $\bm{x}^{traj}$ and force response $\bm{F}^{res}$ in cycles of 1 ms. 
By setting up the optimized reference stiffness beforehand, the optimized position command 
is generated in cycles of 1 ms in response to the contact state as the non-diagonal stiffness 
matrix modifies the trajectory passively in line with the external force. 
Note that the reference stiffness $\bm{K}^{nondiag}$ can be set up 
before the instance of contact because $\bm{K}^{nondiag}$ 
does not have any interference to the trajectory during free motion. 
In contrast, the agent generates the position/force command in cycles of 20 ms 
in the conventional method. 
In summary, while the proposed method ensures the generation of actions corresponding 
to the contact states in cycles of 1 ms, the cycle for generating actions is 20 ms in the conventional method. 
Both methods modify the trajectory, but the responsivity is improved in the proposed method 
due to the shortening of the sampling period of command generation. 

\subsection{Design of stiffness matrices} 
As described in the previous subsection, $\bm{K}^{nondiag}$, the action in the discrete space, 
is designed in advance so that the robot motion is guided in the desired direction. One of the 
advantages of outputting the stiffness matrices as an action of the agent is that the admittance 
model leads to a solid design of stiffness matrices as described below. 

First, the trajectory deviation $\bm{x}^{adm}$ given by the admittance model converges to \begin{eqnarray}
 \bm{x}^{adm}\!\!\!\!\!\! &=& \bm{K}^{nondiag-1}\bm{F}^{res} \label{eq:knondiag-1}\\
  &=& \!\!\!\!\!\!\left[\!\!\begin{array}{cccccc}
		\bm{k}^{inv}_{x} \!\!& \bm{k}^{inv}_{y} \!\!& \bm{k}^{inv}_{z} \!\!& \bm{k}^{inv}_{rx} \!\!& \bm{k}^{inv}_{ry} \!\!& \bm{k}^{inv}_{rz}
	\end{array}\!\!
	\right] \bm{F}^{res}\nonumber	\\
  &=& \!\!\!\!\!\!\left[\!\!\!\!\begin{array}{cccccc}
		k^{inv}_{xx} \!\!& k^{inv}_{xy} \!\!&k^{inv}_{xz} \!\!&k^{inv}_{xrx} \!\!&k^{inv}_{xry} \!\!&k^{inv}_{xrz} \\
		k^{inv}_{yx} \!\!& k^{inv}_{yy} \!\!&k^{inv}_{yz} \!\!&k^{inv}_{yrx} \!\!&k^{inv}_{yry} \!\!&k^{inv}_{yrz} \\
		k^{inv}_{zx} \!\!& k^{inv}_{zy} \!\!&k^{inv}_{zz} \!\!&k^{inv}_{zrx} \!\!&k^{inv}_{zry} \!\!&k^{inv}_{zrz} \\
		k^{inv}_{rxx} \!\!& k^{inv}_{rxy} \!\!&k^{inv}_{rxz} \!\!&k^{inv}_{rxrx} \!\!&k^{inv}_{rxry} \!\!&k^{inv}_{rxrz} \\
		k^{inv}_{ryx} \!\!& k^{inv}_{ryy} \!\!&k^{inv}_{ryz} \!\!&k^{inv}_{ryrx} \!\!&k^{inv}_{ryry} \!\!&k^{inv}_{ryrz} \\
		k^{inv}_{rzx} \!\!& k^{inv}_{rzy} \!\!&k^{inv}_{rzz} \!\!&k^{inv}_{rzrx} \!\!&k^{inv}_{rzry} \!\!&k^{inv}_{rzrz}
	\end{array}\!\!\!\!
	\right]\!\!\!\!\left[\!\!\!\!\begin{array}{c}
		f_x \\ f_y \\ f_z \\ \tau_{x} \\ \tau_{y} \\ \tau_{z} \\
	\end{array}\!\!\!\!
	\right]\nonumber	
\end{eqnarray}
in case the admittance model is stable. This indicates that the local 
trajectory deviation 
$\bm{x}^{adm}$ can be modified by setting up $\bm{K}^{nondiag}$ for an expected 
force $\bm{F}^{res}$. 
Suppose an external force occurs in the $z$ direction, substituting 
$\bm{F}^{res}=[0\ 0\ f_z\ 0\ 0\ 0]^T$ 
into (\ref{eq:knondiag-1}) yields  
\begin{eqnarray}
 \bm{x}^{adm}&=&\bm{k}^{inv}_{z}f_z. 
 \label{eq:kinvzfz}
\end{eqnarray}
To design $\bm{k}^{inv}_z$ for the desired deformation 
$\Delta\bm{x}^{fz}$ against 
the expected force $f^r_z$, (\ref{eq:kinvzfz}) is developed as follows: 
\begin{eqnarray}
 \bm{k}^{inv}_z&=&\Delta\bm{x}^{fz}/f^r_z
\end{eqnarray}
After deriving $\bm{k}^{inv}_x, \bm{k}^{inv}_y, \bm{k}^{inv}_{rx}, \bm{k}^{inv}_{ry}$, 
and $\bm{k}^{inv}_{rz}$ similarly, the stiffness matrix can be calculated using the following formula: 
\begin{eqnarray}
 \!\!\bm{K}^{nondiag}\!\!=\!\!\left[\!\!\begin{array}{cccccc}
		\bm{k}^{inv}_{x} \!\!& \bm{k}^{inv}_{y} \!\!& \bm{k}^{inv}_{z} \!\!& \bm{k}^{inv}_{rx} \!\!& \bm{k}^{inv}_{ry} \!\!& \bm{k}^{inv}_{rz}
	\end{array}\!\!
	\right]^{-1}\!\!\!\!\!\!\!\!
\end{eqnarray}
Although an example of the expected force in only one axis is provided by the above descriptions, 
it can be extended to any direction by multiplying a rotational matrix. 

Multiple types of stiffness matrices 
$\bm{K}^{nondiag}_1,$ $\bm{K}^{nondiag}_2 \cdots \bm{K}^{nondiag}_M $ are derived 
in advance because different actions should be selected to converge the robot motion to a certain 
condition from different states.  

\subsection{Trajectory planning}
\begin{figure}[tb]
    \centering
    \includegraphics[width=7cm]{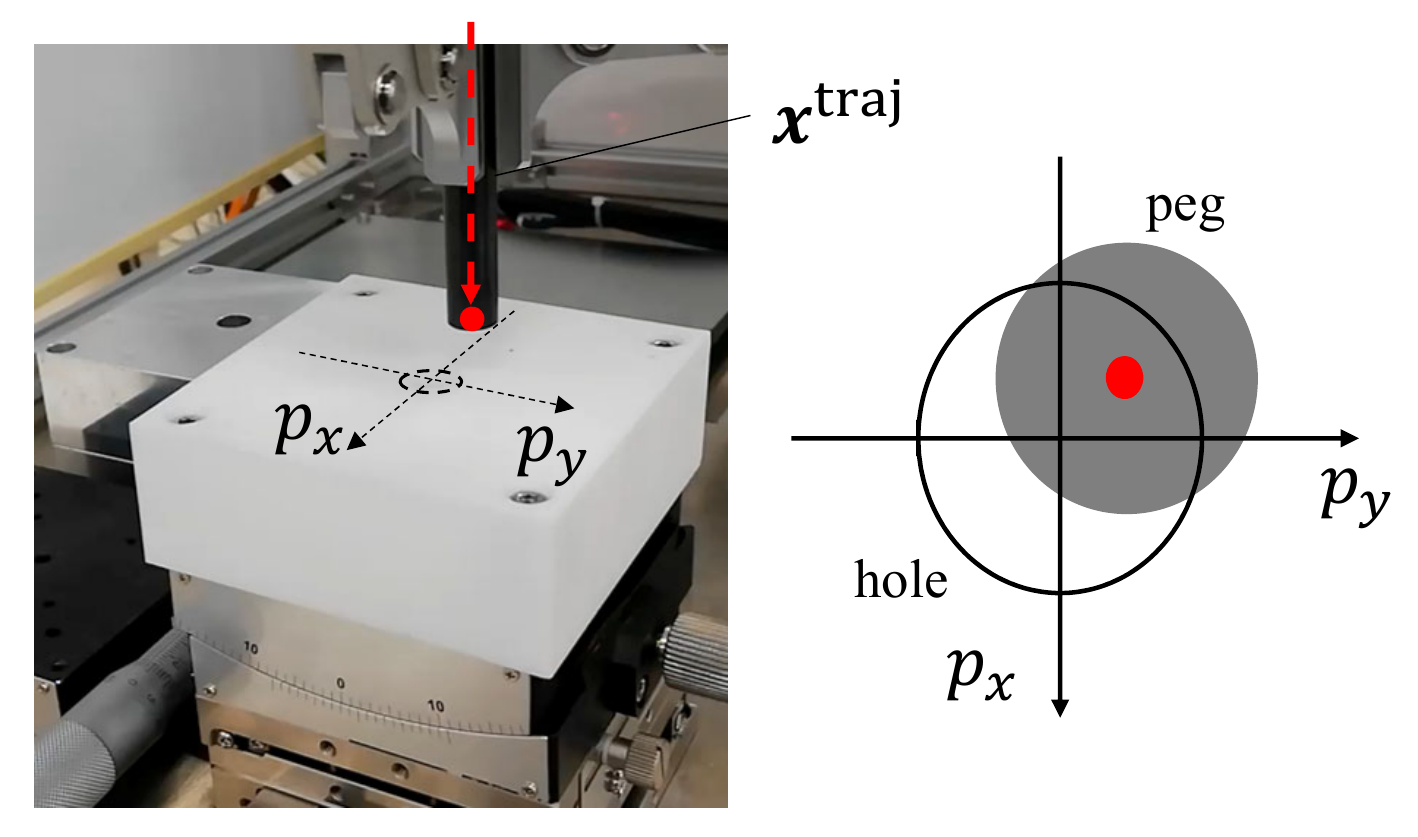}
    \caption{Start and hole position in peg-in-hole task}
    \label{fig:pegin_env}
\end{figure}

As shown in Fig.  \ref{fig:pegin_env}, a simple, pre-determined trajectory is provided, 
moving from the start position downward in the direction of the $z$-axis at constant velocity. 
Moreover, for simplification, the trajectory begins with negative $p_x$ and positive $p_y$ in the experiments. 
The external torque $\tau_x$ and $\tau_y$ is related to $p_x$ and $p_y$ in case the peg and the hole 
are partially overlapped during contact. To detect minute torque through the peg and let the agent estimate the contact state, 
the bottom of the peg and the hole need to be aligned. The preconditions for this peg-in-hole task are as follows.
\begin{itembox}[l]{Preconditions for peg-in-hole task}
  \begin{description}
    \item[1] The hole and the bottom of the peg should overlap or align over a certain value on the $x–y$ plane.
    \item[2] Start with $p_x$ as negative and  $p_y$ as positive in \prettyref{fig:pegin_env}.
  \end{description}
\end{itembox}

\subsection{Q-learning}
RL algorithm starts with a random exploration of the action space $\bm{a}_t$ and strives to 
maximize the cumulative reward $R_t$: 
\begin{eqnarray}
  R_t = r_{t+1} + \gamma r_{t+2} + \gamma^{2} r_{t+2} + \cdots +\gamma^{k} r_{t+k}
  \label{eqn:definition_R}
\end{eqnarray}
where $\gamma$ and $r_t$ are the discount factor and the current reward, respectively. 

Q-learning finds a policy $\pi(\bm{s}_t)$ that maximizes the expected value of the total reward 
over successive steps, starting from the current state as described in the following formula. 
\begin{eqnarray}
   \bm{a}_t = \pi(\bm{s}_{t}) = \argmax_{\bm{a}_t}Q(\bm{s}_{t},\bm{a}_t)
  \label{eqn:Q_Pi}
\end{eqnarray}
Here, the action value function $Q(\bm{s}_{t},\bm{a}_t)$ is updated as follows:
\begin{eqnarray}
\!\!\!\!\!\!\!\! Q(\bm{s}_{t},\bm{a}_t) \!\!\!&\!\!\leftarrow\!\!& \!\!Q(\bm{s}_{t},\bm{a}_t) \nonumber\\
 &&\!\!\!\!\!+\alpha \!\left(\! r_{t}\! + \gamma \maxn_{\bm{a}_{t+1}}Q(\bm{s}_{t+1},\bm{a}_{t+1}) - Q(\bm{s}_{t},\bm{a}_t)\! \right)
  \label{eqn:Q_update}
\end{eqnarray}
where $\alpha$ is a learning rate. .

Deep Q-learning is a method to approximate the action value function $Q$
by a deep neural network (DNN) model. Substituting the parameters of the DNN model $\bm{w}$ 
into (\ref{eqn:Q_update}), the following formula is given. 
\begin{eqnarray}
\!\!\!\!\!\!\!\!\!\!\!\!\!\!\!\!w \!\!\!&\!\!\leftarrow \!\!\!&\!\!w \nonumber\\
 &&\!\!\!\!\!\!\!\!\!\!\!\!+ \alpha\! \left(\! r_{t} \!+\! \gamma \maxn_{\bm{a}_{t+1}}\hat{Q}(\!\bm{s}_{t+1},\bm{a}_{t+1}\!) \!-\! \hat{Q}(\!\bm{s}_{t},\bm{a}_t\!)\! \right) \!\!\frac{\partial \hat{Q}(\!\bm{s}_{t},\bm{a}_t\!)}{\partial w}\!
 \label{eqn:update_param_DQN_new}.
\end{eqnarray}
$\epsilon$-greedy algorithm is introduced to avoid local minimum, 
while $\epsilon$ is gradually reduced with the reduction ratio $\beta$ by 
$\epsilon \leftarrow \beta \epsilon$ . 
The policy $\pi(\bm{s}_{t})$ is described as follows by using $\epsilon$-greedy algorithm
\begin{eqnarray}
    \!\!\!\!\pi(\!\bm{s}_{t}\!) \!= \!\left\{\!\!\! \begin{array}{ll}
        \argmax_{\bm{a}_t}\hat{Q}(\!\bm{s}_{t},\bm{a}_t\!)\! & ,{\tiny{\rm {with\ probability\ }} (\!1\!-\!\epsilon\!)\!\!\!} \!\!\!\! \\
        {\rm random}\! & ,{{\rm \tiny{with\ probability\ }} \normalsize{\epsilon}} \\
        \end{array} \right. \!\!\!\!
\end{eqnarray}

\section{Verification experiment}
\subsection{Verification system set-up}
\begin{figure}[tb]
    \center
    \includegraphics[width=6cm]{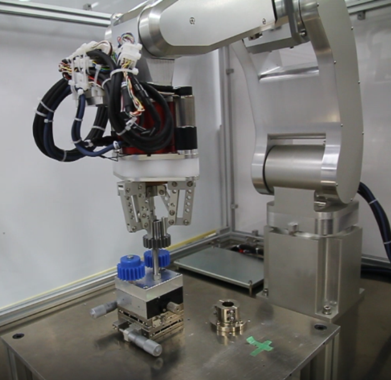}
    \caption{Robot arm used in this study}
    \label{fig:robot}
\end{figure}

\begin{figure}[tb]
    \centering
    \includegraphics[width=7cm]{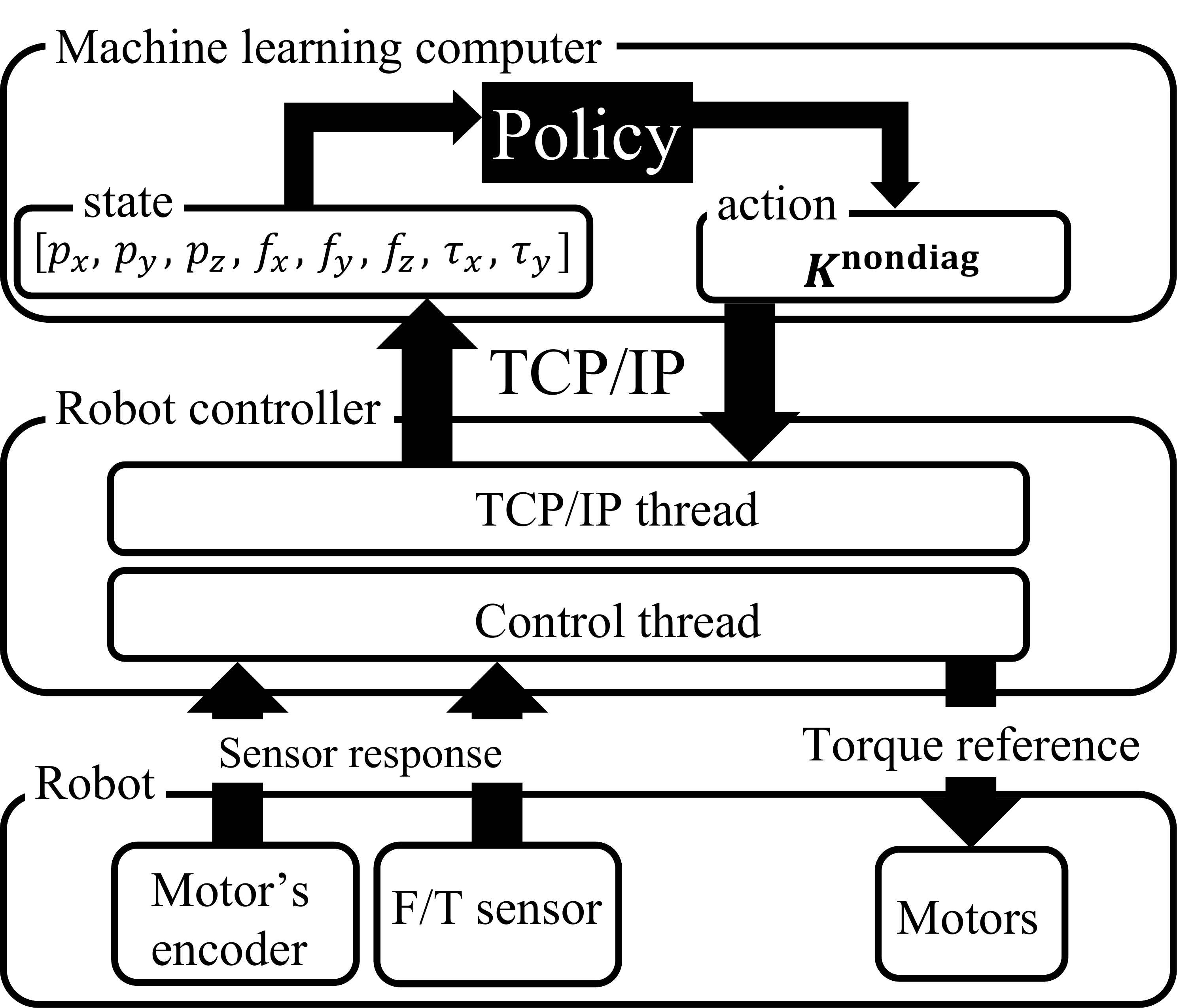}
    \caption{Configuration of the proposed system}
    \label{fig:system_exp}
\end{figure}

Fig. \ref{fig:robot} shows the six-DOF robot manipulator used in the experiment. 
A three-finger gripper was fixed on a six-axis force sensor, and the net force and 
moment applied on the grasped object was measured. Fig. \ref{fig:system_exp} is 
an overview of the proposed system. TCP/IP communication is used to transmit 
NN input data between the robot's control computer and the NN control computer. 

\subsection{Comparison of control performance}
First, control performances with different sampling time of admittance model calculation 
were compared with a peg-in-hole task. Since admittance model generates the position 
reference of local trajectory optimization, the result with 20 ms sampling time should be 
similar to that of a conventional RL algorithm with position command output. By shortening 
the sampling time, the deviation of force response during contact reduced. Additionally, 
the recognition time of contact, the time interval between the contact time and the 
time the agent selected the action for contact, was also shortened owing to smaller 
deviation of force information. The result in 20 ms sampling time ended in an 
unstable motion and it implies that the conventional methods requires more conservative 
control setup for contact motion. In sum, shortening the sampling time of the admittance 
model is essential for performance improvement of contact tasks.    
\begin{table}[htb]
    \centering
    \caption{Control performance with different sampling time of admittance model}
    \begin{tabular}{l||l|l|l} \hline
	Sampling  & Recognition time  & Max. deviation & Av. deviation\\
	time & of contact & during contact & during contact\\
	\hline
	1ms  & 736 ms & 5.30 N & 1.44 N\\
	2ms  & 1122 ms & 7.72 N & 2.40 N\\
	5ms  & 1188 ms & 13.0 N & 4.94 N \\
	20ms  & N/A & N/A & N/A \\
	\hline
    \end{tabular}
\end{table}
\subsection{Peg-in-hole task}
The diameters of the peg and hole are 10.05 and 10.07 mm, respectively, leaving a gap of 20 $\mu$m. For the peg-in-hole task, a stiffness matrix is used where the interference is only in the translational direction. The matrix in the rotational direction has a constant value as below.

\begin{eqnarray}
  \bm{K}^{r}
  =\left[
   \begin{array}{ccc}
    50 & 0 & 0 \\
    0 & 50 & 0 \\
    0 & 0 & 50
  \end{array}
  \right]
\end{eqnarray}

In the peg-in-hole task, the agent accepts the state shown in (\ref{eqn:DQN_state}) and, as the action 
defined in (\ref{eqn:DQN_action}), chooses from the following four types of non-diagonal stiffness matrices. \\
\begin{eqnarray}
  &1)& \bm{K}^{nondiag}_1=\left[\begin{array}{cccc}
     525 & 194 & -194 & \\
	194 & 662 & 137 & \bm{0}\\
	-194 & 137 & 662 & \\
	 & 0 & & \bm{K}^r 
    \end{array}\right]\nonumber
\end{eqnarray}

\begin{eqnarray}
  &2)& \bm{K}^{nondiag}_2=\left[\begin{array}{cccc}
     525 & 0 & -275 & \\
	0 & 800 & 0 & \bm{0} \\
	-275 & 0 & 525 & \\
	 & \bm{0} & & \bm{K}^r 
    \end{array}\right]\nonumber\\
  &3)& \bm{K}^{nondiag}_3=\left[\begin{array}{cccc}
     800 & 0 & 0 & \\
	0 & 525 & 275 & \bm{0} \\
	0 & 275 & 525 & \\
	 & \bm{0} & & \bm{K}^r 
    \end{array}\right]\nonumber\\
  &4)& \bm{K}^{nondiag}_4=\left[\begin{array}{cccc}
     300 & 0 & 0 & \\
	0 & 300 & 0 & \bm{0} \\
	0 & 0 & 800 & \\
	 & \bm{0} & & \bm{K}^r 
    \end{array}\right]\nonumber
\end{eqnarray}

\begin{figure}[tb]
    \centering
    \includegraphics[width=6cm]{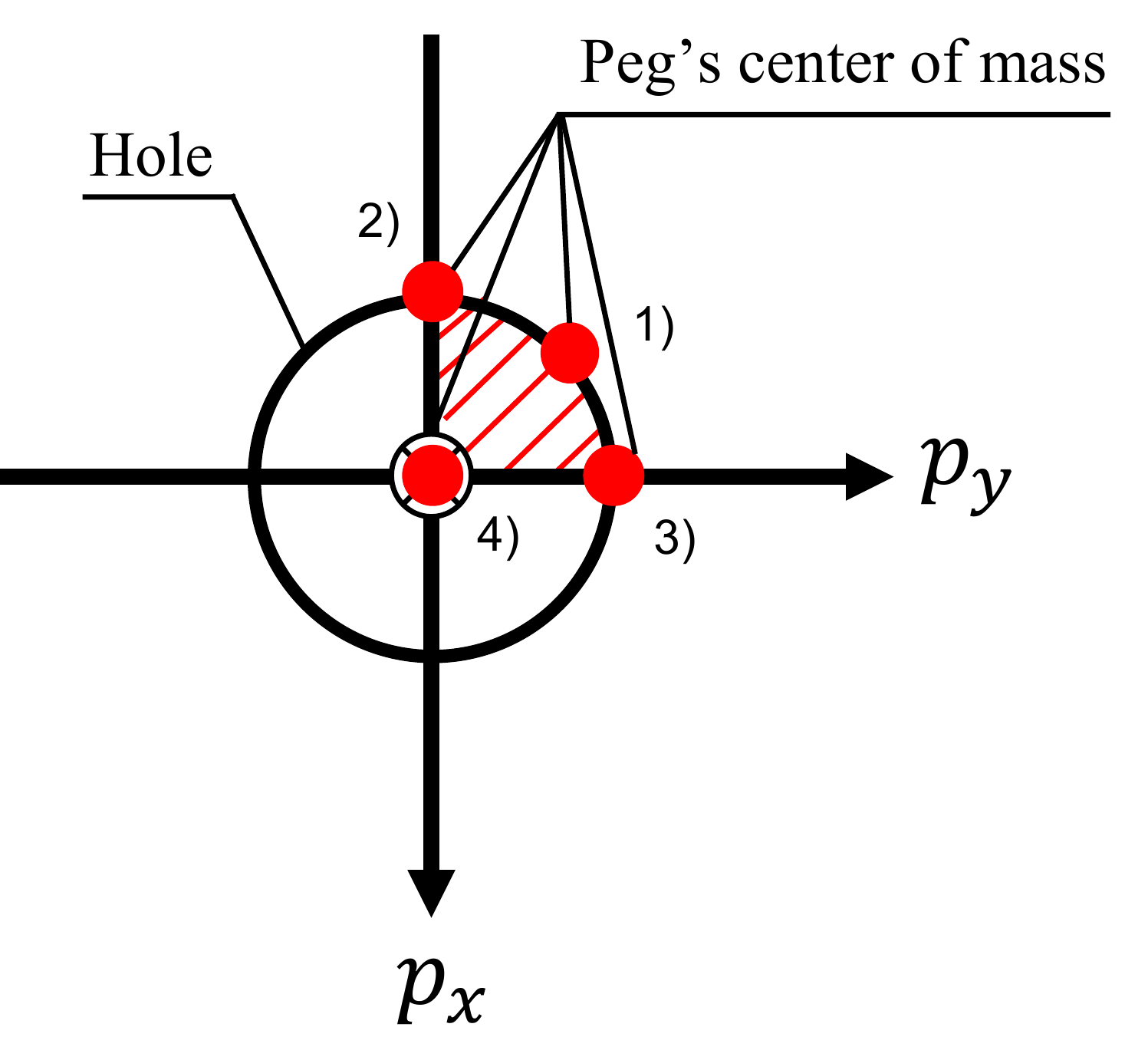}
    \caption{Ideal selection of stiffness matrices in peg-in-hole task}
    \label{fig:peg_stiff_select}
\end{figure}

Considering the preconditions for this task, the stiffness matrices can be narrowed down to 
four, as shown in Fig. \ref{fig:peg_stiff_select}. 
When the peg is at location 1), $\bm{K}^{nondiag}_1$ should be seelcted as the action and 
the displacement in the direction of the hole is generated by the admittance model. 
Suppose 1N contact force on the z-axis was generated, the admittance model modifies the trajectory 
for +5 mm and -5 mm in $x$ and $y$-axes, respectively.  
When the peg is at location 2) or 3), $\bm{K}^{nondiag}_2$ or $\bm{K}^{nondiag}_3$ should be 
selected as the action and the displacement is generated in $x$ and $y$ direction, respectively. 
The peg being at location 4) constitutes the insertion phase; thus, the task can be performed 
robustly in relation to any angle disturbance by decreasing the stiffness in the $x, y$ direction 
and increasing stiffness in the $z$ direction. The state accepted by the agent is redefined as follows.
(\ref{eqn:DQN_state}) has been normalized because the amplitude of the force and torque is different to each other. 
\begin{eqnarray}
\bm{s}_t=\left[p_x, p_y, \frac{p_z - p_z^{i}}{p_z^{i} - p_z^{g}}, \frac{f_x}{f_x^{m}},  \frac{f_y}{f_y^{m}},  \frac{f_z}{f_z^{m}},  \frac{\tau_{x}}{\tau_{x}^{m}}, \frac{\tau_{x}}{\tau_{y}^{m}} \right] \!\!\!\!\!\!
 \label{eqn:state_exp}
\end{eqnarray}
where $i, g, $ and $m$ in superscripts denote initial, goal, and maximum values,respectively.  
$f_x^{m}, f_y^{m}, f_z^{m}, \tau_x^{m}$ and $\tau_y^{m}$ were set as 20 N, 20 N, 40N, 1 Nm, and 1 Nm, respectively. 

The reward $r_t$ was designed as follows:  

\begin{eqnarray}
    r_{sparce} = \left\{ \begin{array}{ll}
        1-\frac{t}{K} & ,{\rm task \ finished}  \\ \nonumber
        & \\ \nonumber
        -1 &            ,{\rm task \ failed}   \nonumber
        \end{array} \right.
\end{eqnarray}
$K$ is the number of maximum steps, which was 500 in this study.

\begin{figure}[tb]
    \centering
    \includegraphics[width=8cm]{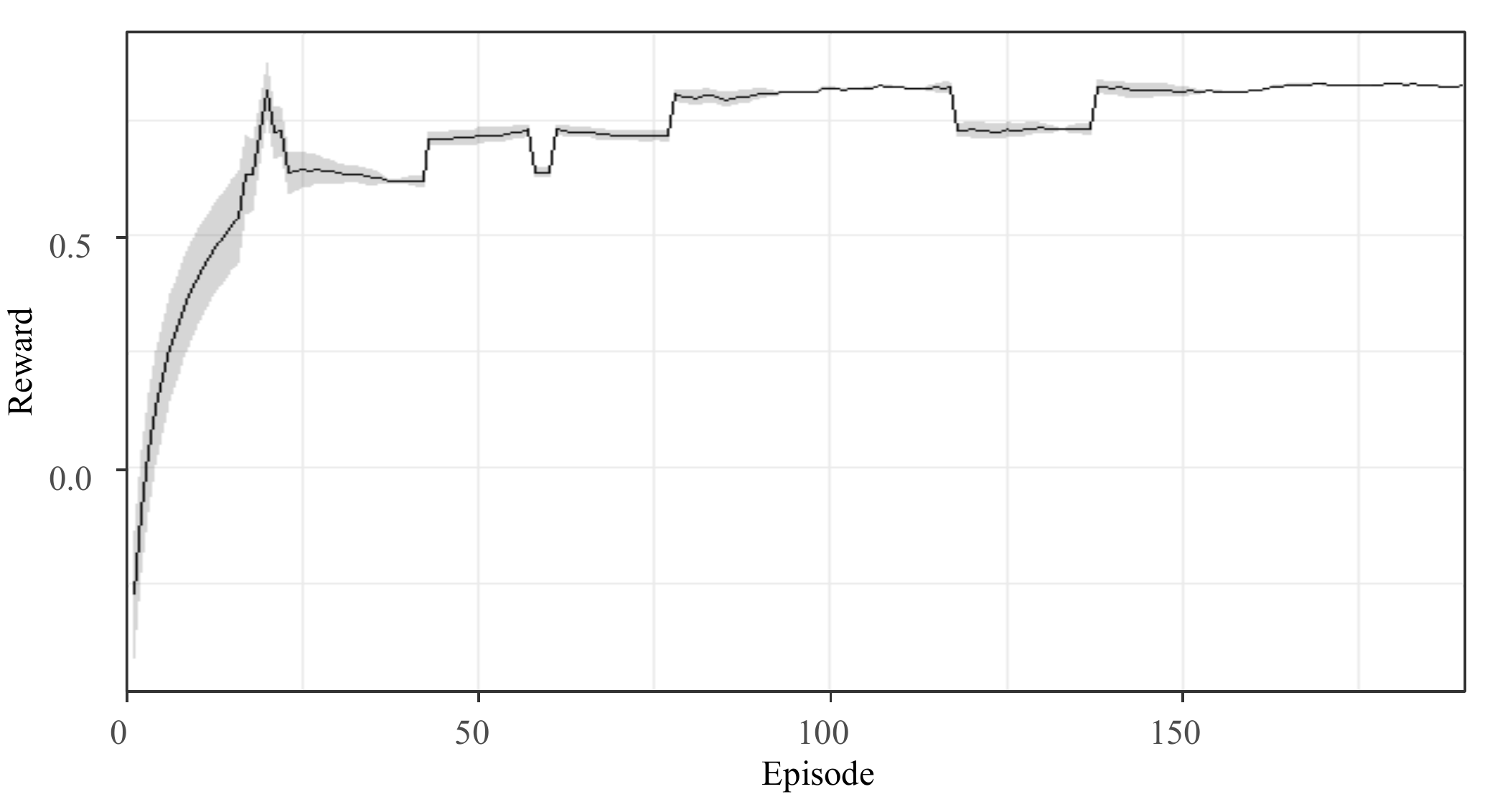}
    \caption{Reward during learning}
    \label{fig:reward}
\end{figure}

Fig.~\ref{fig:reward} shows the learning progress in the case of a 20$\mu$m clearance with 0 degrees 
tilt angle, and 3mm initial offset 
Means in a moving window of 20 episodes are shown in a solid line and their 90\% confidence interval 
is layered as the gray area. It shows that the reward converged in about 150 episodes. 

\begin{figure}[tb]
    \centering
    \subfloat[$p_x$]{
        \includegraphics[width=4cm]{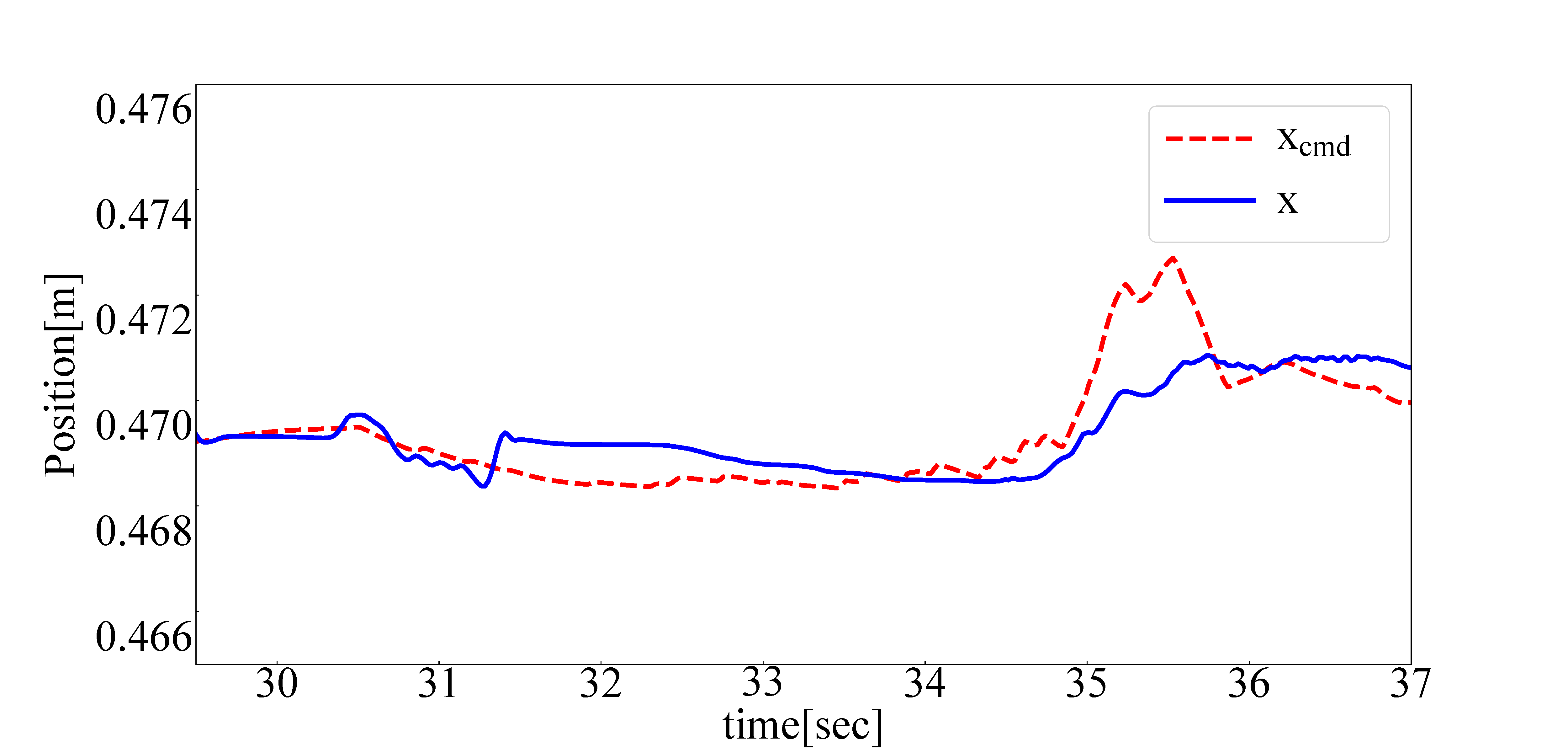}
        \label{fig:pegin_pos_force_response/px}}
    \subfloat[$p_y$]{
        \includegraphics[width=4cm]{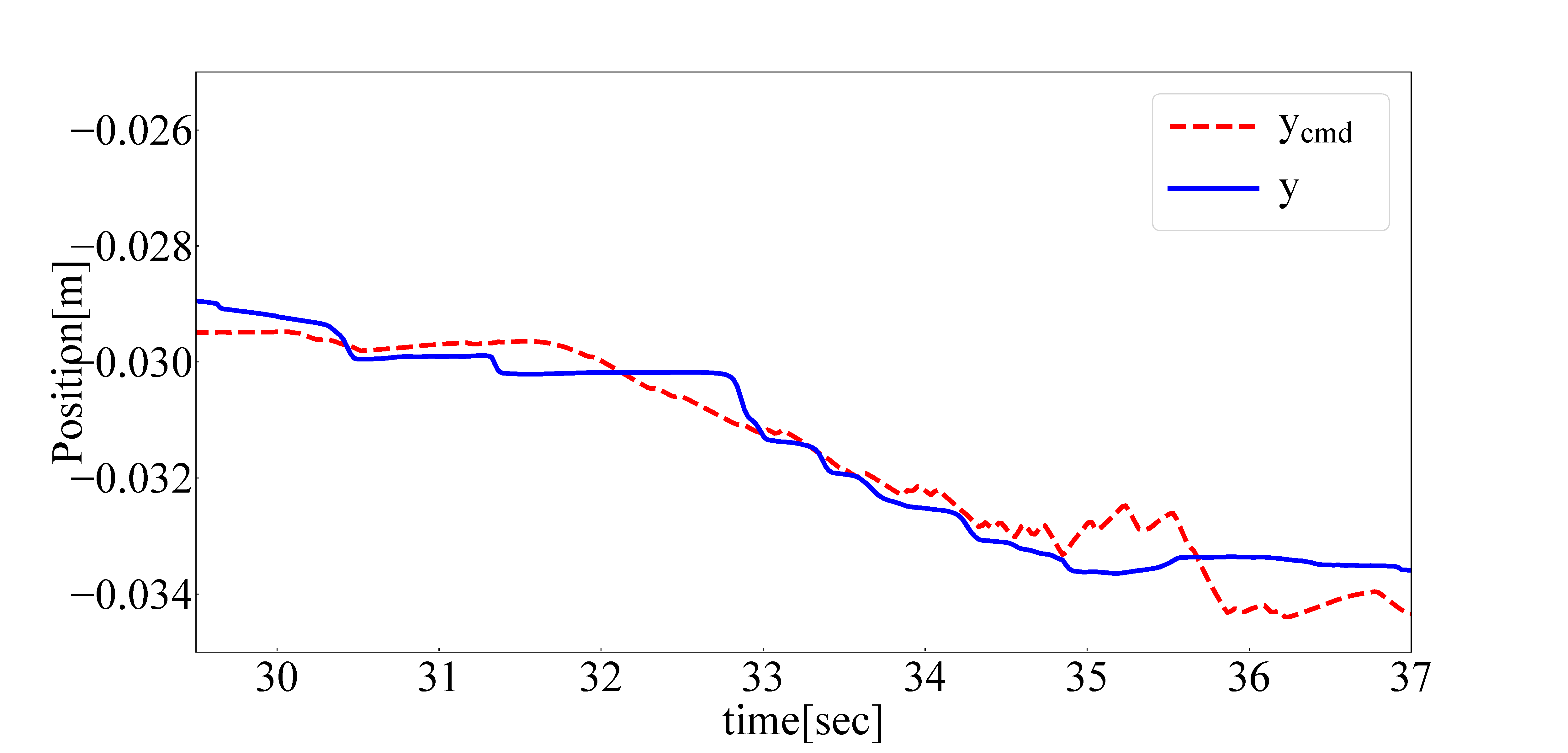}
        \label{fig:pegin_pos_force_response/py}}\\
    \subfloat[$p_z$]{
        \includegraphics[width=4cm]{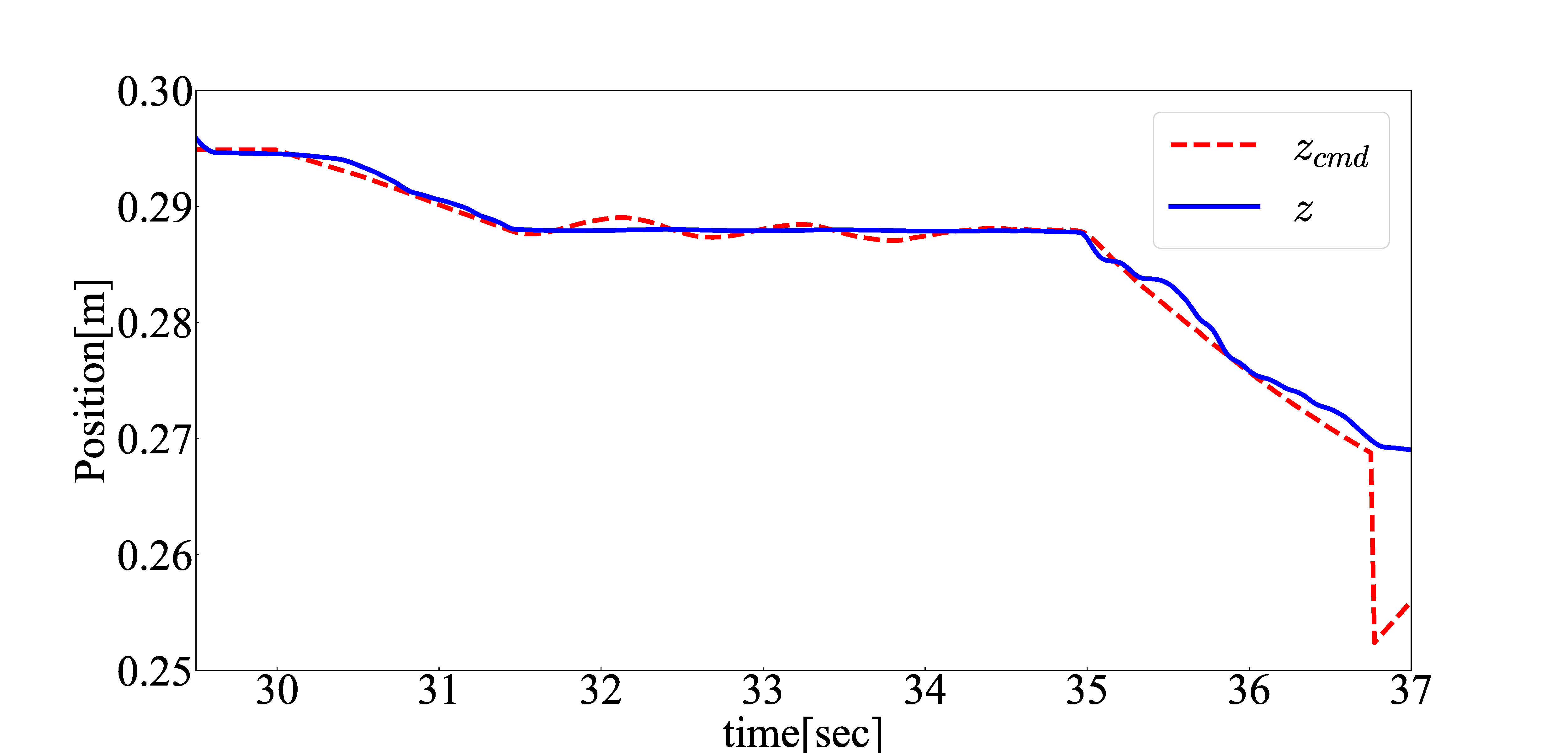}
        \label{fig:pegin_pos_force_response/pz}}
    \subfloat[$f_x$]{
        \includegraphics[width=4cm]{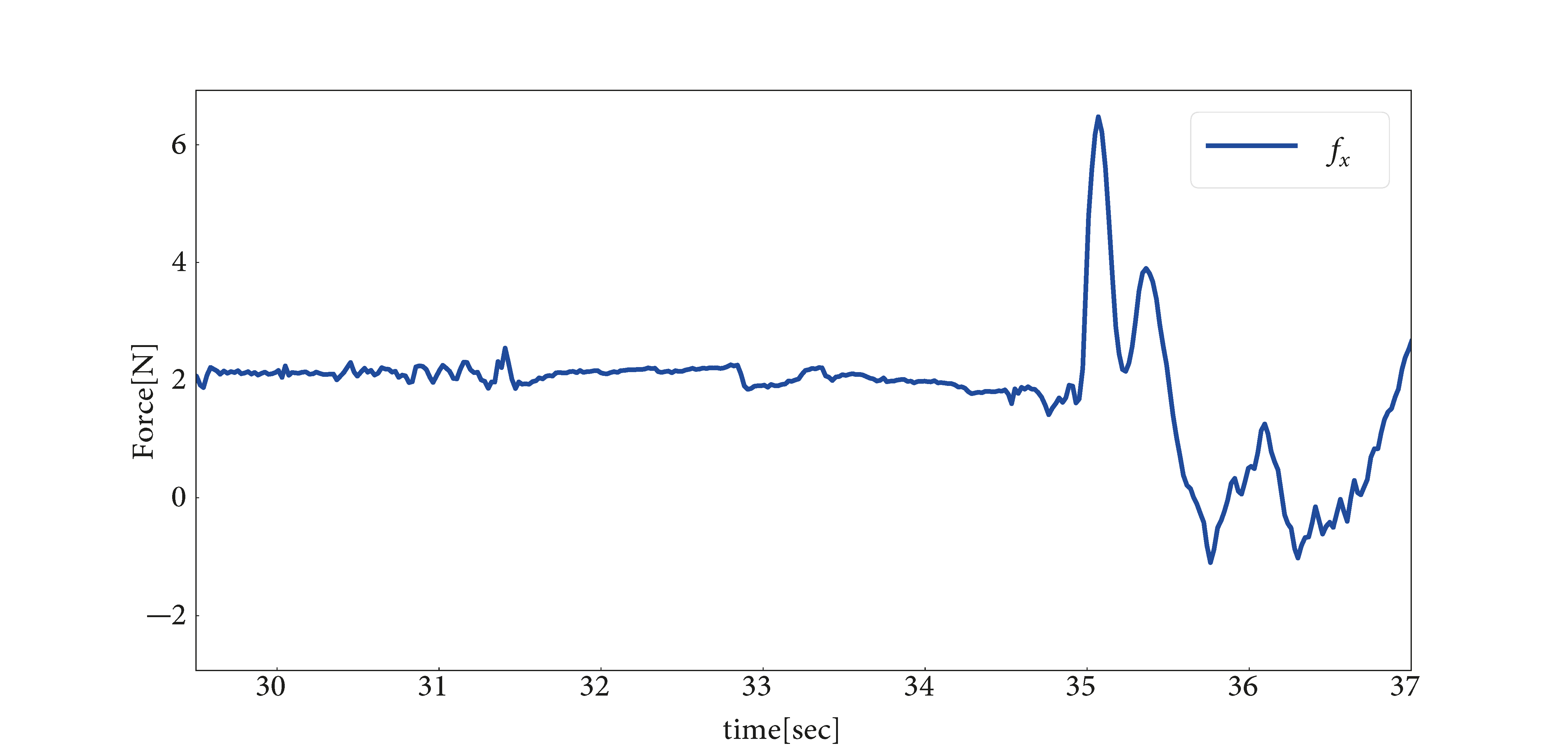}
        \label{fig:pegin_pos_force_response/fx}}\\
    \subfloat[$f_y$]{
        \includegraphics[width=4cm]{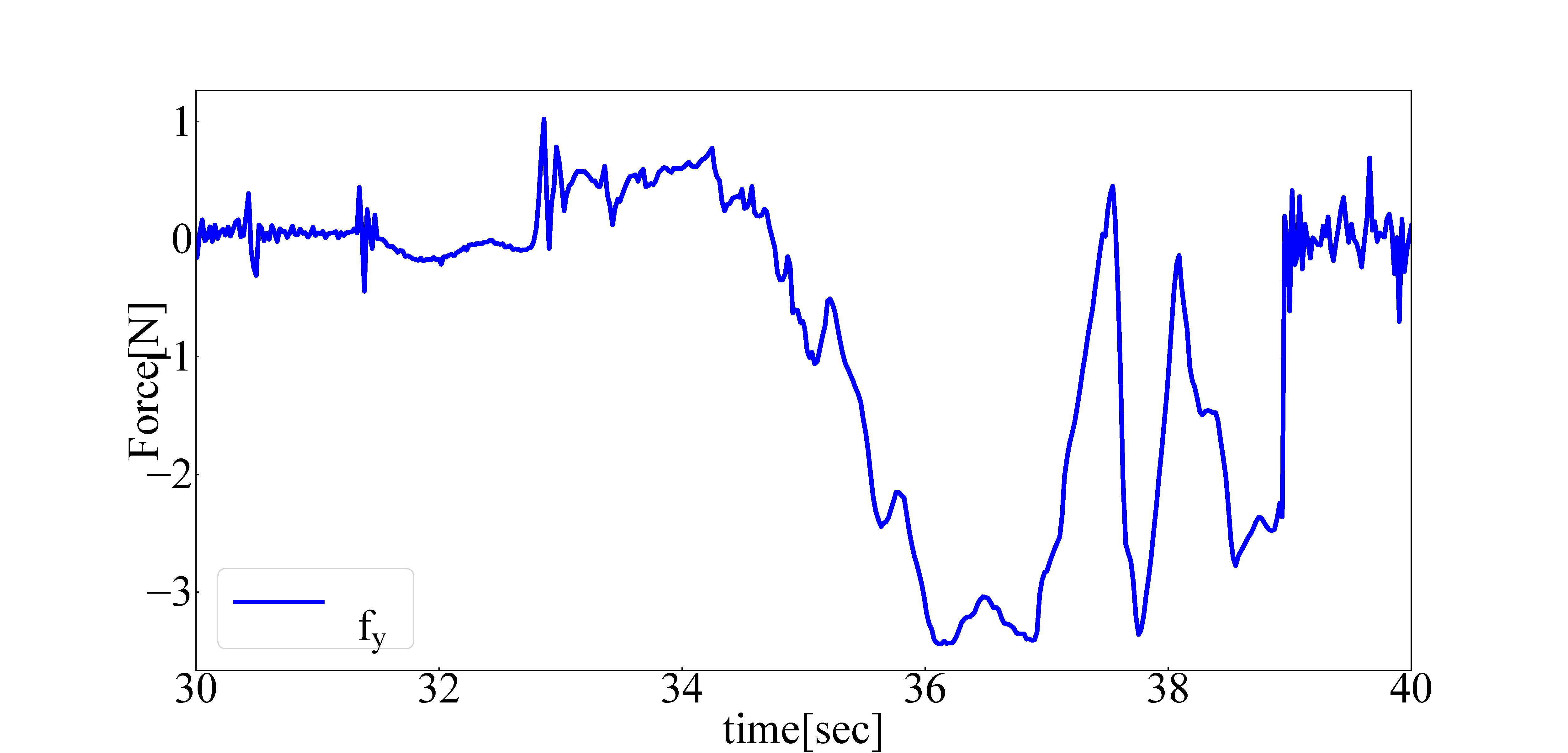}
        \label{fig:pegin_pos_force_response/fy}}
    \subfloat[$f_z$]{
        \includegraphics[width=4cm]{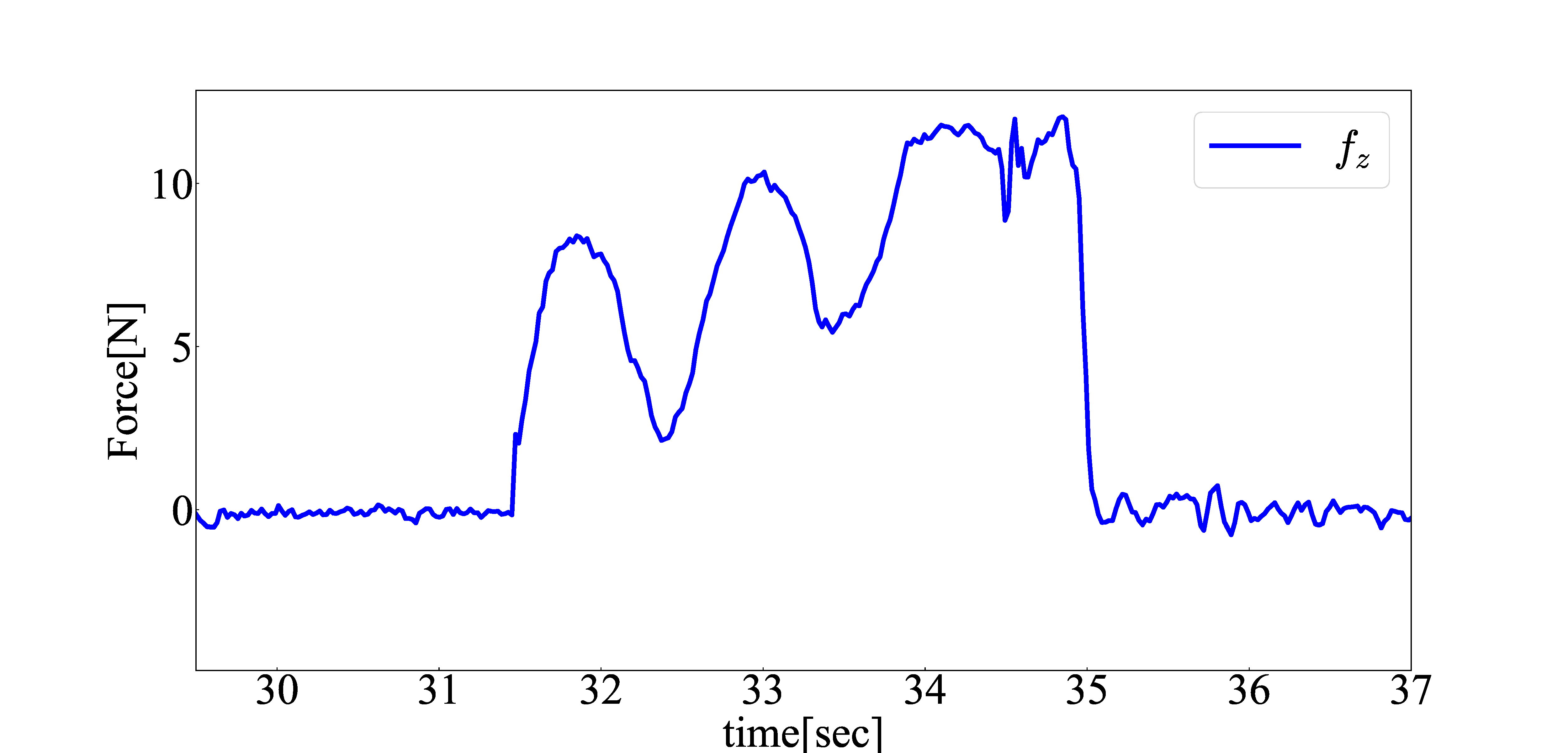}
        \label{fig:pegin_pos_force_response/fz}}\\
    \subfloat[$\tau_{x}$]{
        \includegraphics[width=4cm]{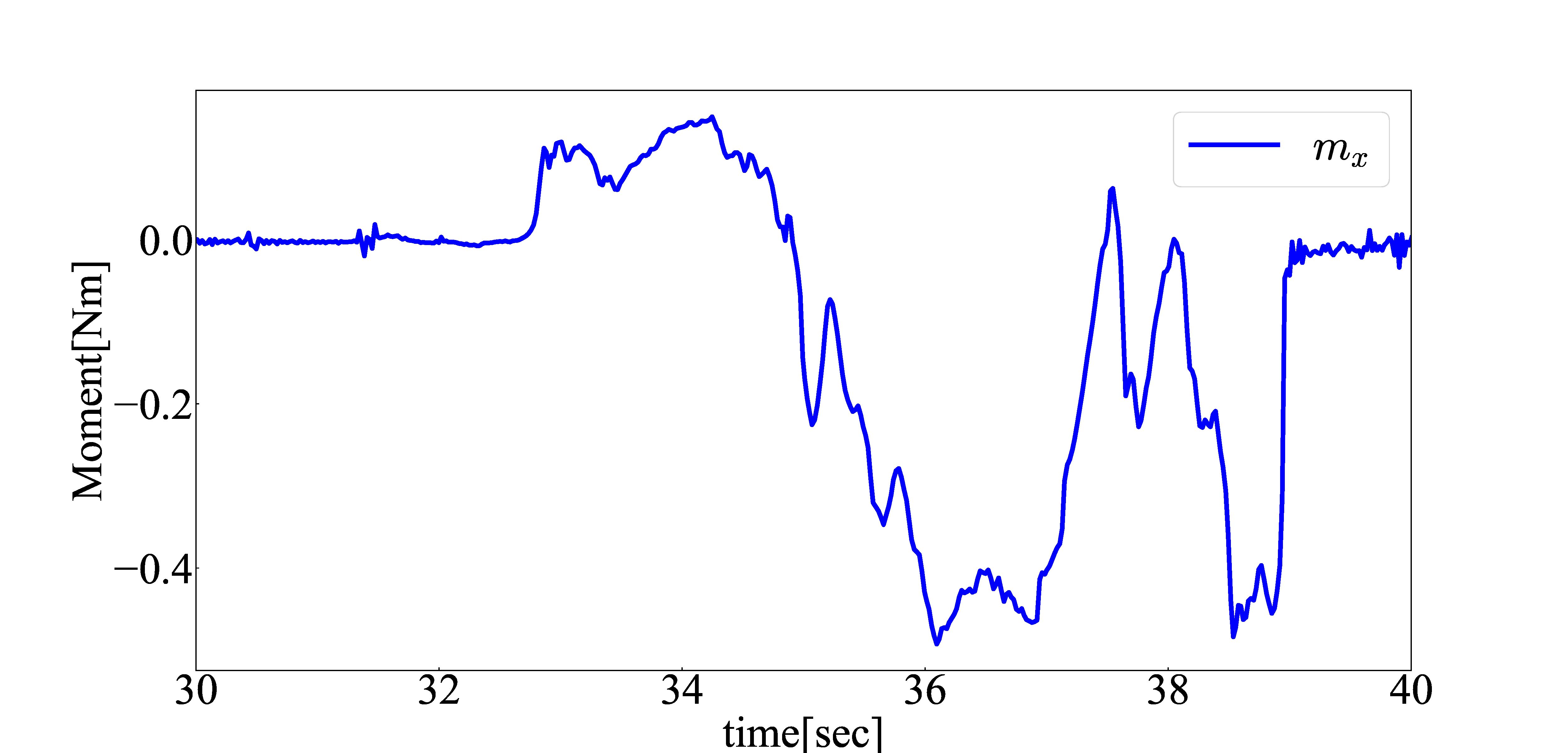}
        \label{fig:pegin_pos_force_response/mx}}
    \subfloat[$\tau_{y}$]{
        \includegraphics[width=4cm]{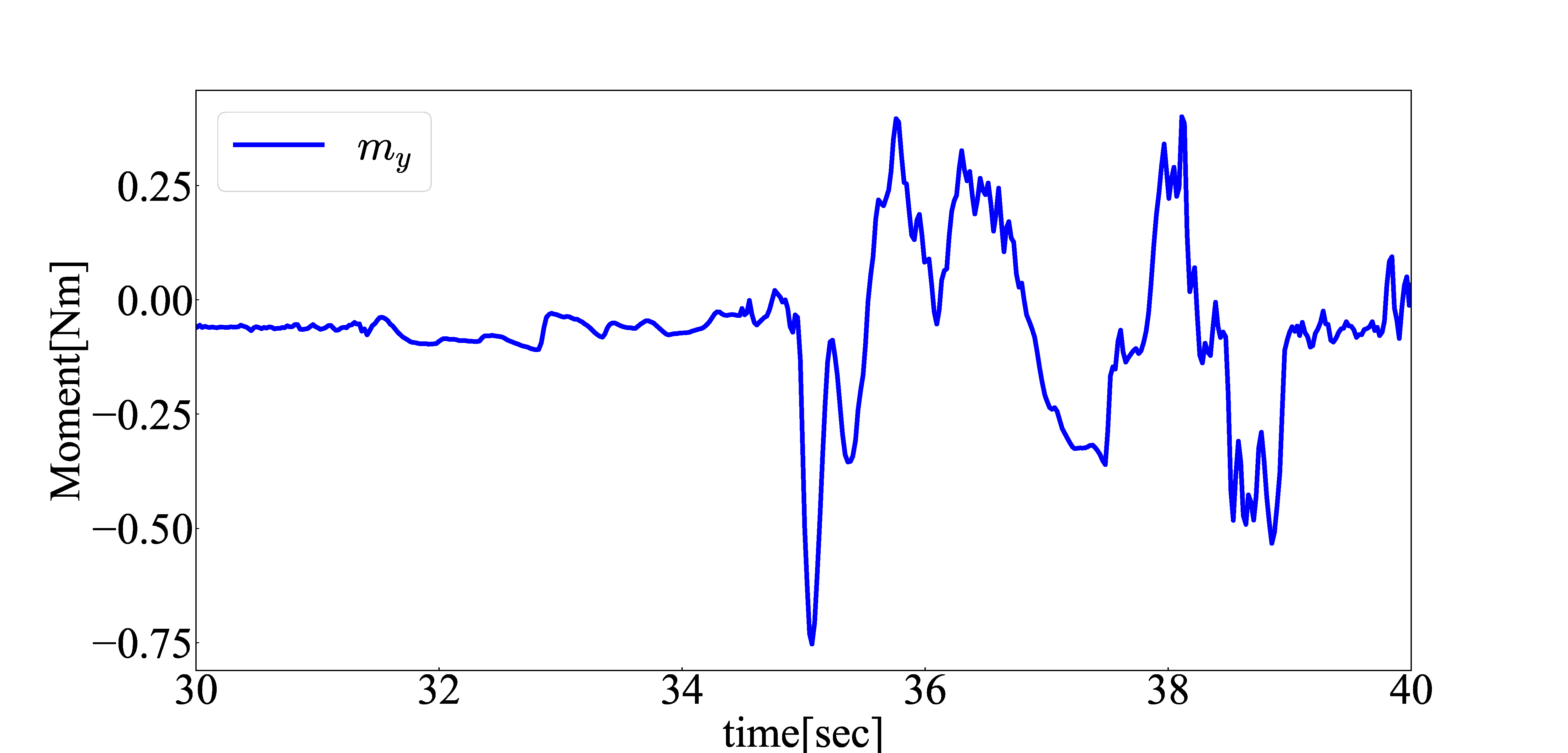}
        \label{fig:pegin_pos_force_response/my}}
    \caption{Position and force responses during a peg-in-hole task}
    \label{fig:pegin_pos_force_response}
\end{figure}

\begin{figure}[tb]
    \centering
    \includegraphics[width=7cm]{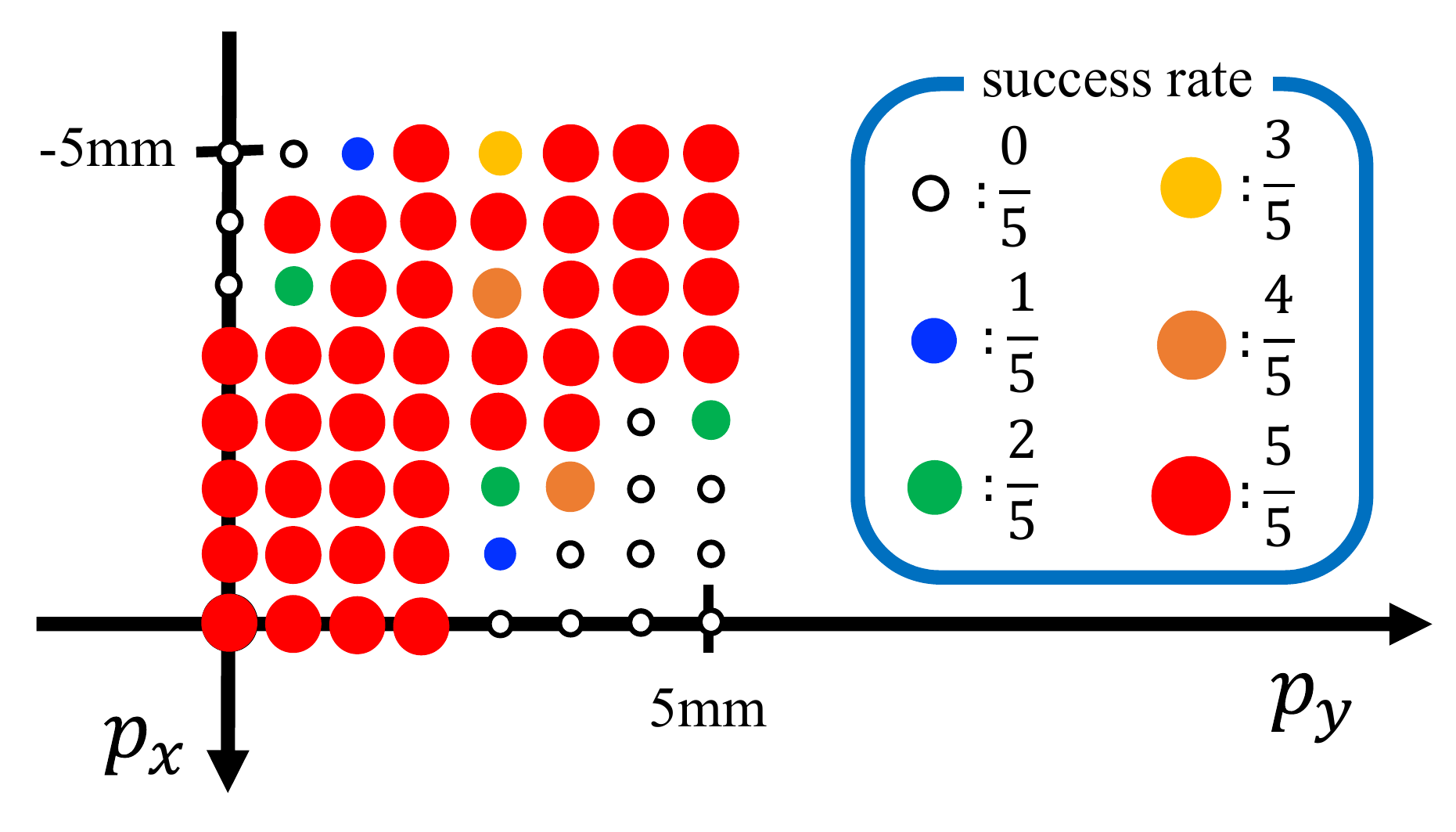}
    \caption{Success rate of peg-in-hole task with different position error}
    \label{fig:pegin_success_rate}
\end{figure}

Fig. \ref{fig:pegin_pos_force_response} 
shows the position on task execution and the response to force values. The maximum absolute 
force and the moment applied on the finger are $6$ N and $0.75$ Nm, respectively, which are 
much smaller than a reaction force that causes damage to the parts. 
Fig.~\ref{fig:pegin_success_rate} shows the success rates of the task with different initial position errors. 
Since the position accuracy of the robot is not smaller than 1mm, various initial position errors 
were produced by adjusting the initial position of the hole by a manual x-y linear stage. 
When the position errors for the $x$– and $y$–axes were both smaller than 3 mm, the success 
rate was 100\%. On the other hand, the success rate was low when the position error was larger 
than 3 mm in one of the axes and small in the other axis. Since there were too small overlap between 
the peg and the hole, the agent could not find a proper action. When the error was larger in both 
axes, the success rate was still high although the precondition 1 was not met. Since 
$\bm{K}_1^{nondiag}$ was selected as an action when no overlap between the peg and the hole 
exists, the peg moved toward the direction of the hole and as a result, precondition 1 was met 
finally. 

\begin{figure}[tb]
    \centering
    \includegraphics[width=7cm]{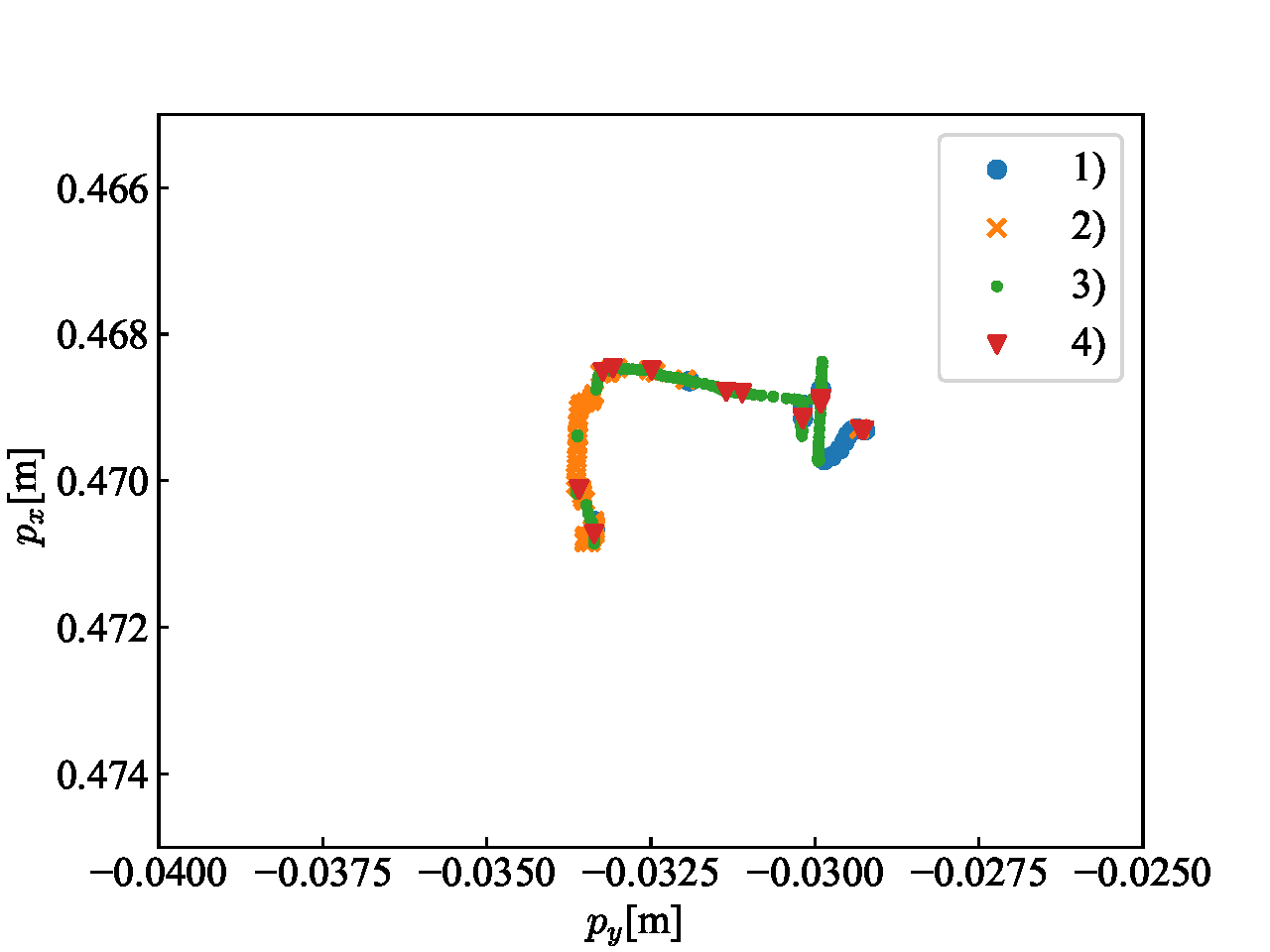}
    \caption{Selected actions during peg-in-hole task}
    \label{fig:pegin_exp_dist}
\end{figure}

Fig. \ref{fig:pegin_exp_dist} shows the actions selected during a learning phase of a peg-in-hole task. 
Some samples show that improper stiffness matrices were selected at times because of 
$\epsilon$-greedy algorithm. This phenomena can be eliminated during a test phase by eliminating 
the $\epsilon$-greedy algorithm. 
Other samples shows that $\bm{K}^{nondiag}_3$ was mainly selected at the early stage when $p_y$ is larger 
and $\bm{K}^{nondiag}_2$ was selected when  $p_y$ was around -0.033 m. $\bm{K}^{nondiag}_4$ 
was selected when the peg center was located in the center of the hole. 
Note that the exact position of the hole is unknown because the robot has few millimeters position 
error in general. Hence,the results infers that force/torque information had a strong influence to  
the agent for selecting the stiffness matrices.  Additionally, 
it is evident from Fig. \ref{fig:pegin_exp_dist} that the agent was 
trained to ensure that the stiffness matrices guide the peg to the direction which the hole exists. 

\begin{figure}[tb]
    \centering
    \includegraphics[width=8.4cm]{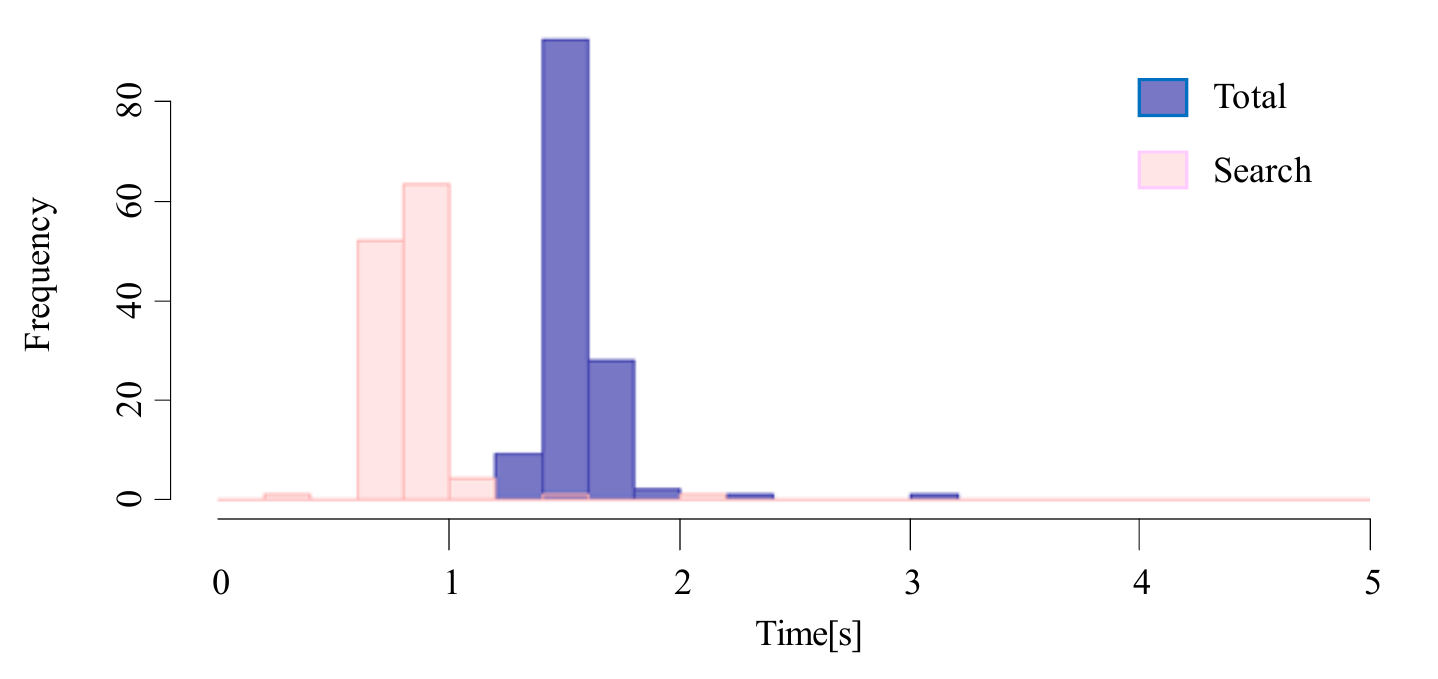}\\
	\footnotesize {(a) (-3mm, 3mm) initial offsets, 0 degree tilt}
    \includegraphics[width=8.4cm]{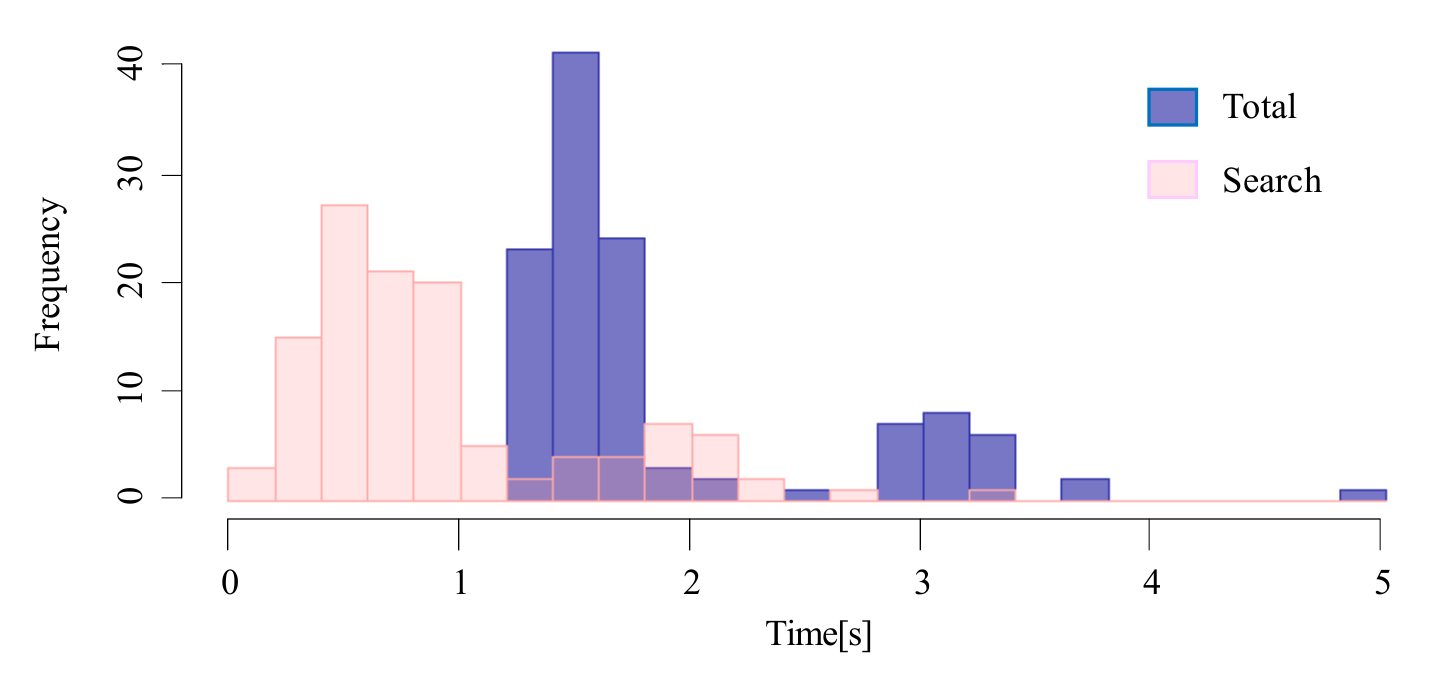}
	\footnotesize{(b) (-3mm, 3mm) initial offsets, 2 degree tilt}
    \caption{Histogram of execution time for peg-in-hole task}
    \label{fig:histogram_pegin}
\end{figure}

To show the robustness and the rapid responses of the proposed method, we performed 
experiments with pegs of 20 $\mu$m clearances. 
The performances can be seen in https://youtu.be/gxSCl7Tp4-0 
by a video.  
The peg-in-hole task was executed for 100 times after 
learning to see the time for the peg-in-hole motion. We examined the two cases: (a) (-3mm, 3mm) initial offsets in the x-y plane, and 0 degree tilted angle, (b) (-3mm, 3mm) initial offsets in the x-y plane,  
and 2 degree tilted angle. Fig.~\ref{fig:histogram_pegin} shows the distribution of the execution 
time using histograms. The results show that search time and the insertion time have been shortened
with the proposed method in both the cases. The average total time was 1.64 s and 1.87 s, 
respectively, which are less than 50\% compared to the previous state of the art study~\cite{inoue2017deep}. 

\begin{figure}[tb]
    \centering
    \includegraphics[width=8cm]{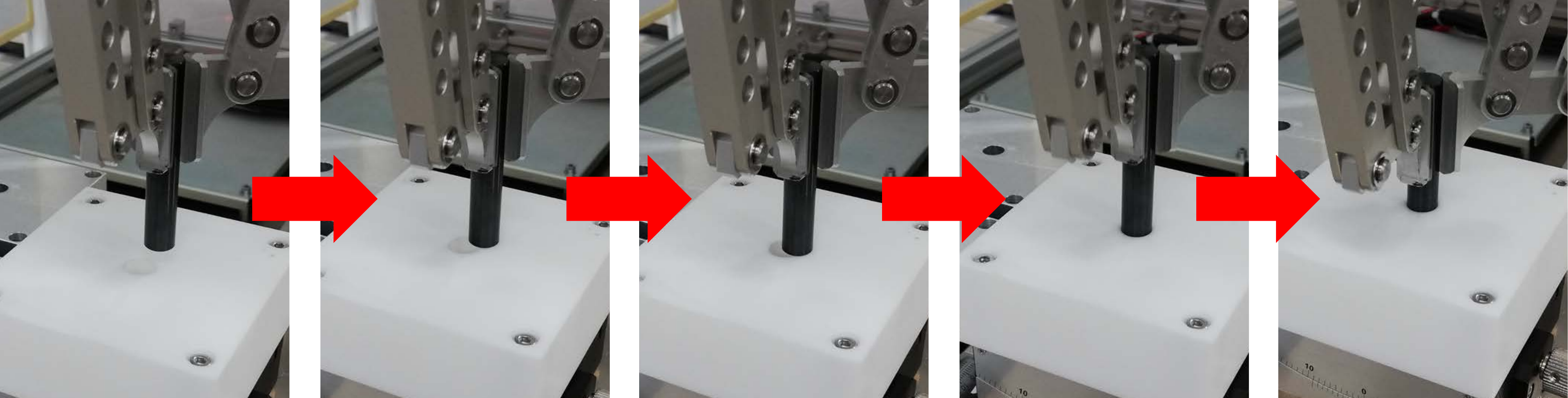}
    \caption{Snapshots of a peg-in-hole task}
    \label{fig:pegin_exp_pic}
\end{figure}
\subsection{Gear-insertion task}
\begin{figure}[tb]
    \centering
    \includegraphics[width=8.4cm]{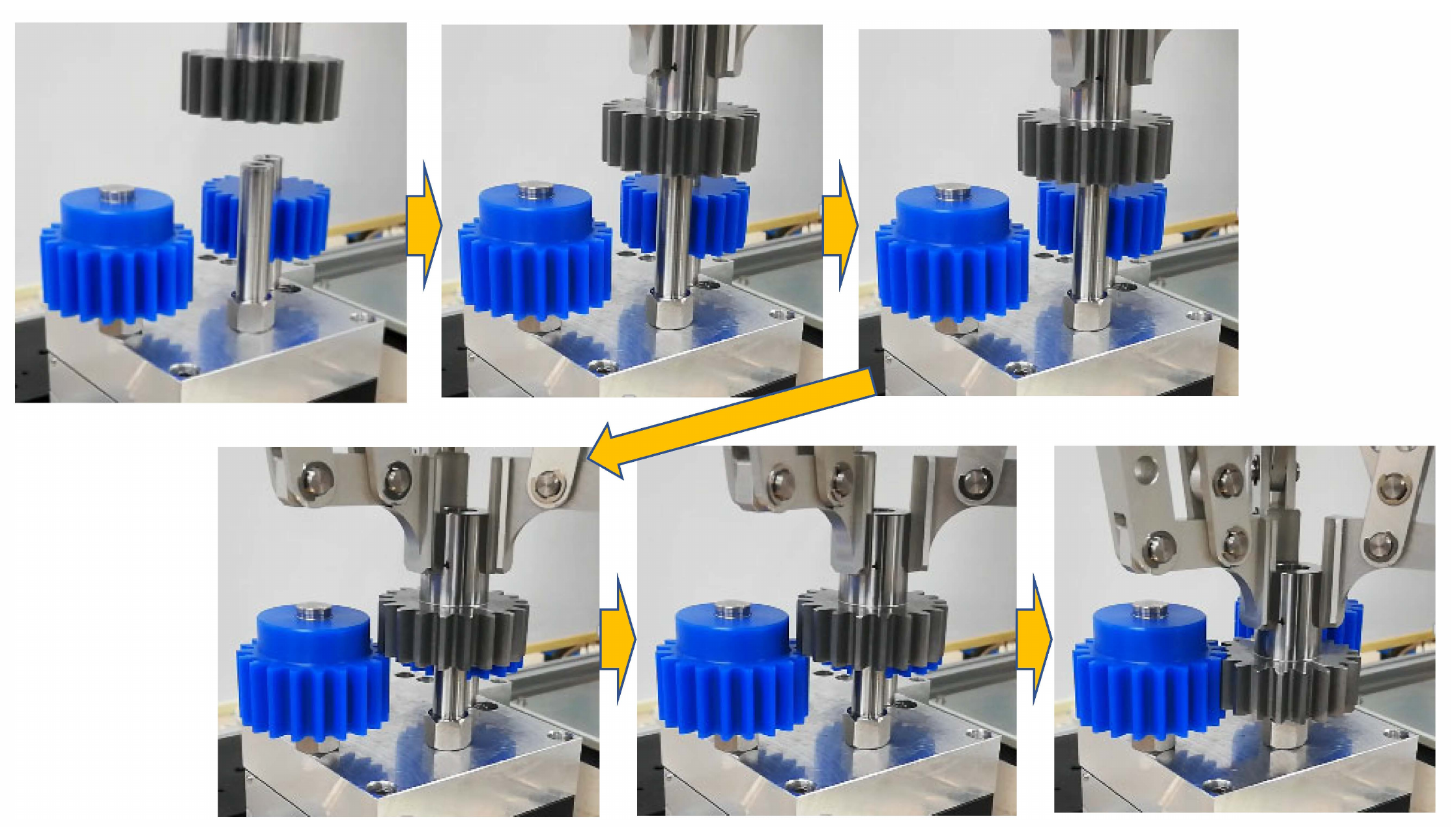}
    \caption{Snapshots of gear-insertion task}
    \label{fig:gearinsertionsnap}
\end{figure}

\begin{figure}[tb]
    \centering
    \includegraphics[width=8.4cm]{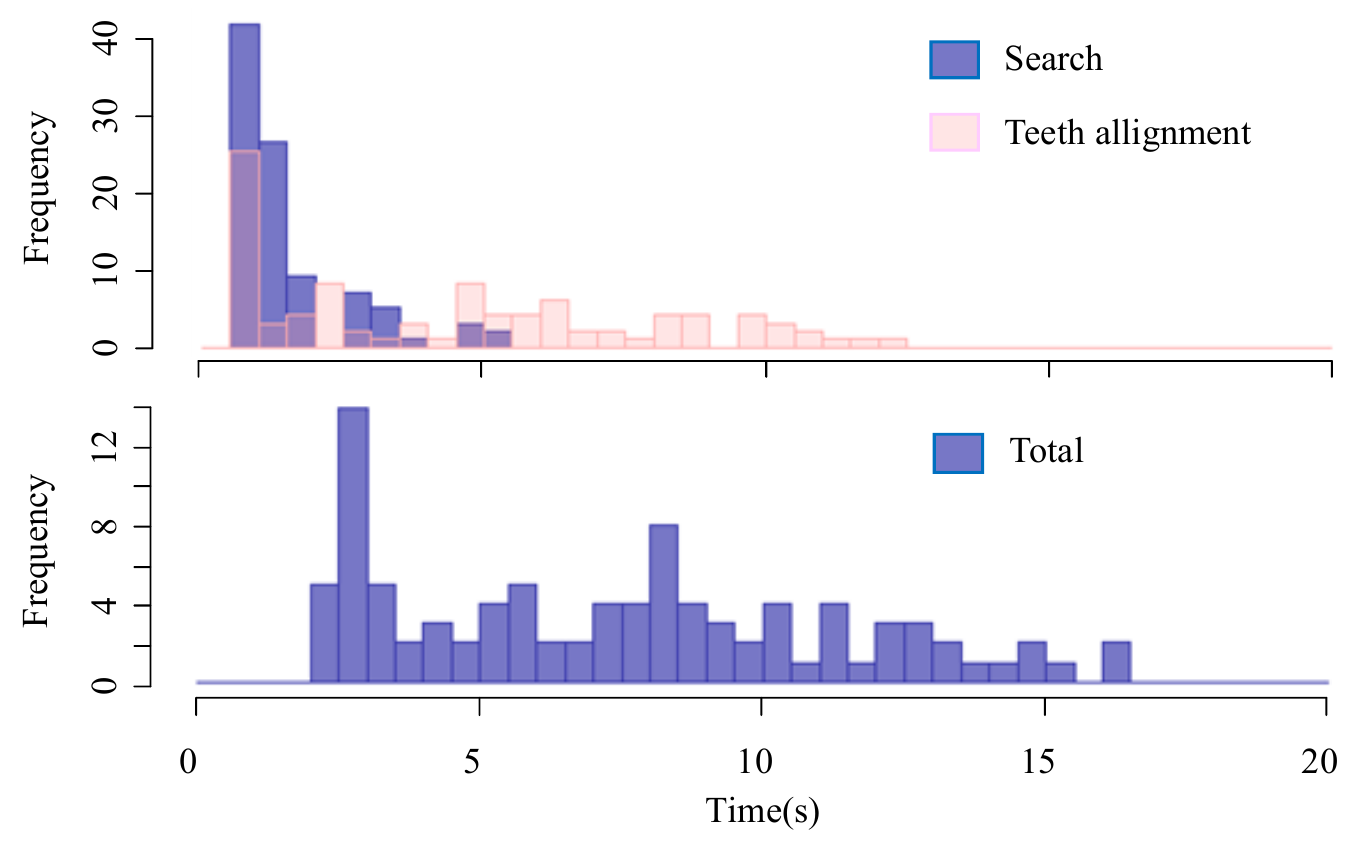}
    \caption{Histogram of execution time for gear-insertion task}
    \label{fig:histogram_gear}
\end{figure}

For the gear-insertion task, a similar condition with the peg-in-hole task: 
20 $\mu$m clearance and a 10 mm radius peg, was introduced. 
The four non-diagonal stiffness matrices 
$\bm{K}_1^{nondiag}, \cdots, \bm{K}_4^{nondiag}$ were also used 
in the action space. 
However, the robot needed to align the gear teeth with that of two 
pre-fixed gears. Therefore, another non-diagonal stiffness matrix was 
added to the action space as follows:  
 \begin{description}
  \item[5)]  $\bm{K}^{nondiag}_5=\left[\begin{array}{cccc}
     300 & 0 & 0 & \\
	0 & 300 & 0 & \bm{0} \\
	0 & 0 & 400 & \\
	 0 & 0 & 0 & \\
	 0 & 0 & 0 & \bm{K}^r \\
	 0 & 0 & 200\sin(\omega t) &   
    \end{array}\right]$
\end{description}
Since the angles of the two pre-fixed gears were unknown, 
Gear-teeth alignment was done by rotating around z-axis 
by a sinusoidal wave after insertion. Fig.~\ref{fig:gearinsertionsnap} 
shows the snapshots of an experiment. Commercial spur gears with module 2 
for gear box assembly was used in this study to evaluate the performance 
in a practical standard. Fig.~\ref{fig:histogram_gear} 
shows the histogram of the execution time. The results show that the 
execution time strongly depends on the teeth alignment time, which 
stochastically varies. One noticeable point of this experiment is that 
a contact-rich task with several contact transitions was accomplished 
with a simple linear trajectory with variable admittance model. 
The gear-insertion task in the video   
(see https://youtu.be/gxSCl7Tp4-0 )
shows that the grasped 
gear moved toward the peg center after contact and also rotated 
around z-axis after insertion, although it only generates a linear 
trajectory without any contact.   
Another interesting point is that the time distribution of the searching 
time was similar to that of Fig.~\ref{fig:histogram_pegin} despite 
the difference of inserting a grasped peg into a hole and inserting a 
fixed peg into a grasped part. The results shows that these two 
can be treated with a similar control architecture. 

\section{Conclusion}
This study proposed a method for the generation of non-diagonal stiffness matrices online for 
admittance control using deep Q-learning. The actions were generated corresponding to the 
contact states, according to the agent's update cycles for contact-rich tasks using conventional 
machine learning. However, the proposed method ensures the local trajectory optimization in line with 
the contact states according to the robot's control cycle. The responsiveness and the robustness 
were evaluated through experiments on a peg-in-hole task. and a gear-insertion task with different 
conditions. 
This study handled a contact-rich task in an approach involving a simple trajectory and online 
generation of stiffness matrices. As the proposed method allows for parallel connection 
with a trajectory planning module, it is expected to be applied to a variety of contact-rich manipulations.

\end{document}